\documentclass{article}
\usepackage[margin=1.0in]{geometry}

\usepackage[utf8]{inputenc} 
\usepackage[T1]{fontenc}    
\usepackage{hyperref}       
\usepackage{url}            
\usepackage{booktabs}       
\usepackage{amsfonts}       
\usepackage{nicefrac}       
\usepackage{microtype}      

\usepackage{authblk}

\usepackage{amsmath}
\usepackage{dsfont} 
\usepackage{graphicx}
\usepackage{caption}
\usepackage{subcaption}
\usepackage{float} 
\usepackage{tabularx}
\usepackage{longtable}
\usepackage{arydshln}
\usepackage{multirow} 
\usepackage[table]{xcolor}
\usepackage{pdflscape} 
\usepackage{flushend}
\usepackage{algorithm}
\usepackage{algorithmic}
\usepackage{tikz}
\usetikzlibrary{bayesnet} 

\usepackage[round,numbers,sort&compress]{natbib}

\newcommand{\ie}{i.e., }

\title{A generative, predictive model for menstrual cycle lengths that accounts for potential self-tracking artifacts in mobile health data}
\author[
		1,2]{Kathy Li}
\author[
		1,2]{I\~nigo Urteaga}
\author[3]{Amanda Shea}
\author[3,4]{Virginia J. Vitzthum}
\author[1,2]{Chris H.~Wiggins}
\author[*,5,2]{No\'emie Elhadad}

\affil[1]{Department of Applied Physics and Applied Mathematics, Columbia University, New York NY 10027}
\affil[2]{Data Science Institute, Columbia University, New York NY 10027}
\affil[3]{Clue by BioWink, Adalbertstra{\ss}e 7-8, 10999 Berlin, Germany}
\affil[4]{Kinsey Institute \& Department of Anthropology, Indiana University, Bloomington IN 47405}
\affil[5]{Department of Biomedical Informatics, Columbia University, New York NY 10032}

\affil[*]{Corresponding author: No\'emie Elhadad (noemie.elhadad@columbia.edu)}

\begin{document}
\maketitle


\begin{abstract}
Mobile health (mHealth) apps 
such as
menstrual trackers provide a rich source of self-tracked health observations that can be leveraged for health-relevant research. However, such data streams have questionable reliability since they hinge 
on
user adherence to 
the app. Therefore, it is crucial for researchers to separate true behavior from self-tracking artifacts. By taking a machine learning approach to modeling self-tracked cycle lengths, we can both make more informed predictions and learn the underlying structure of the observed data.
In this work, we propose and evaluate a hierarchical, generative model for predicting next cycle length based on previously-tracked cycle lengths that accounts explicitly for the possibility of users skipping tracking their period. Our model offers several advantages: 1) accounting explicitly for self-tracking artifacts yields better prediction accuracy as likelihood of skipping increases; 2) because it is a generative model, predictions can be updated online as a given cycle evolves, and we can gain interpretable insight into how these predictions change over time; and 3) 
its hierarchical nature enables modeling of
an individual's cycle length history while incorporating population-level information. 
Our experiments using mHealth cycle length data encompassing over 186,000 menstruators with over 2 million natural menstrual cycles show that our method yields state-of-the-art performance against neural network-based and summary statistic-based baselines, while providing insights on disentangling menstrual patterns from self-tracking artifacts. This work can benefit users, mHealth app developers, and researchers in better understanding cycle patterns and user adherence.
\end{abstract}


\section{Introduction}
\label{sec:intro}
Menstruation has been historically misunderstood and understudied, despite its importance to women's health; it impacts women's emotional, physical, and mental well-being, influencing reproductive ability, cardiovascular health, and the presence of chronic diseases~\cite{popat_menstrual_2008, bedford_prospective_2010, zittermann_physiologic_2000, solomon_menstrual_2002,j-Carmina1999, shuster_premature_2010}. Note that while in this paper we may refer to menstruators as `women,' we acknowledge that not all menstruators are women and vice versa. Many menstruators experience the adverse effects of menstruation at some point in their lifetimes -- for instance, dysmenorrhea, or painful menstruation characterized by symptoms such as abdominal cramps and headaches, is estimated to affect up to 91\% of women of reproductive age~\cite{10.1093/epirev/mxt009}. Studies on risk factors and comorbidities of dysmenorrhea have shown its association with quality of life symptoms like depression, anxiety, decreased productivity, and fatigue~\cite{upsala, evans}. Certain menstruation-related disorders are associated with dysmenorrhea, including polycystic ovary syndrome (PCOS) and endometriosis, which are characterized by symptoms like infertility, intense pelvic pain, decreased mobility, and depression~\cite{10.1093/humupd/dmt027, moradi}.
Better understanding of the day-to-day patterns of menstruation can empower women and healthcare professionals to identify and manage such conditions.

Regardless of the role that menstruation plays in women's lives, it has been consistently overlooked as an important area of research, due to a variety of factors, including societal stigma, normalization of women's pain, lack of knowledge of uterine and menstrual physiology, and limited access to reliable datasets~\cite{CRITCHLEY2020624, ASSANIE201986}. 
Gender bias in medicine has resulted in the systemic neglect of conditions that disproportionately affect women, leaving them with misdiagnoses and unaddressed pain~\cite{doing_harm}.
Improved understanding of physiological processes like menstruation can aid in closing these gaps~\cite{vitzthum}. 
The vast difference in understanding and solutions for male and female health conditions, despite the fact that menstruation affects half of the world's population, demonstrates how far the field of menstrual research has yet to go.

The varied, fluctuating nature of the individual menstrual experience can make it difficult to understand and characterize. Traditionally, it has been primarily studied through survey results; studies have shown that experiences vary both within and between individuals, encompassing not only period length and cycle length (the number of days between subsequent periods~\cite{j-Vitzthum2009}), but also qualitative symptoms like period flow, physical pain, and quality of life characteristics ~\cite{j-Arey1939,j-Treloar1967, j-Chiazze1968,j-Muenster1992,j-Belsey1997,j-Burkhart1999,j-Vitzthum2000,j-Creinin2004,j-Williams2006,j-Vitzthum2009,cole}. 
Efforts have also been made to model and predict menstrual cycle lengths, including representing menstrual cycle lengths as a combination of standard and nonstandard cycles with mixture modeling~\cite{10.1093/biostatistics/kxi043} and linear random effects models~\cite{HARLOW19911015}, accounting for the fact that menstrual cycle behavior evolves with age~\cite{HARLOW2000722}; linear mixed models that include clustering to account for between-women variance~\cite{10.2307/2533552}; and state-space modeling that accounts for between-women and within-women variation~\cite{10.1093/biostatistics/kxq020}. 

More recently, efforts have been made to describe menstrual cycles and symptoms~\cite{pierson2021daily,j-Bull2019,j-Li2020}, as well as related physiological events like ovulation~\cite{j-Symul2019, soumpasis} in a quantitative way using self-tracking mobile health (mHealth) data from apps. Such data provide a new opportunity to investigate menstruation on a large-scale, but also require special consideration due to their user-tracked nature. 
While existing studies importantly acknowledge the necessity of modeling menstrual variability~\cite{10.1093/biostatistics/kxi043, HARLOW19911015,HARLOW2000722,10.2307/2533552,10.1093/biostatistics/kxq020}, there is also a need to understand the underlying mechanisms of menstruation and to develop models that account for the specific nature of mHealth data. mHealth data has opened opportunities to better understand a diverse range of individuals across geographies -- we can phenotype various behaviors like physical activity and stress~\cite{j-Althoff2017, j-Smets2018}, as well as characterize diseases like endometriosis, schizophrenia, and Parkinson's~\cite{bot2016mpower,j-Torous2018,ip-Urteaga2018, urteaga2020}. The rise of popularity in menstrual trackers, which are the second most popular app for adolescent girls and the fourth most popular for adult women~\cite{j-Wartella2016,Fox2012}, has in particular enabled access to large-scale, longitudinal cycle length data for rigorous inquiry into how to best characterize menstruation ~\cite{ip-Pierson2018,j-Symul2019,j-Bull2019,j-Li2020}. However, while such mHealth data opens the possibilities of menstrual research, it can be prone to inconsistent adherence from users. Therefore, the difficulty of modeling menstruation holds especially true for such self-tracking data -- researchers must not only take into consideration the inherent variability of users' menstrual experience, but their tracking behavior as well. In this paper, we argue that a generative machine learning model allows us to 
distinguish true menstrual cycle behavior from these self-tracking artifacts.

We take a machine learning approach to modeling menstrual cycle lengths: we infer underlying cycle dynamics and make predictions for future cycle lengths. Probabilistic models, including generative models, have proven useful for healthcare research, as they allow researchers to explicitly represent dependencies across observed and unobserved variables, consider missing values, and characterize model uncertainty~\cite{Chen2020ProbabilisticML}. In particular, we propose a generative model, a statistical framework that defines how data (in this case, cycle lengths) are generated and specifies the relationships between different variables of interest (i.e., cycle lengths and self-tracking adherence).

This provides interpretability, because in addition to learning how to predict menstrual cycle lengths accurately, we also learn the underlying probability distributions that characterize them. It also provides flexibility, because we can explicitly encode variables that we want to represent in the data generation process, such as the presence of potential cycle length skips. In particular, our model encodes a measure of per-user ``typical'' cycle length (i.e., cycle length regularity) and parameterizes explicitly a per-user propensity to skip tracking (i.e., self-tracking adherence) separately. By explicitly accounting for potential tracking artifacts, we can alert users when they are likely to have misstracked a cycle in their history, which has the potential to improve the efficacy of self-reporting. In addition, our model is hierarchical, meaning that we can represent different levels of information about our population. Namely, we can learn shared cycle length characteristics that describe the whole population of users, as well as individual-specific patterns describing each user's behavior. This hierarchy allows us to represent individual-level variability, while also accounting for behaviors and patterns that describe the broader population. 

By using cycle length information only (users most commonly utilize menstrual health apps for period tracking), we attain valuable insights into cycle dynamics and user adherence, ultimately delivering accurate cycle length predictions. We train our model on self-tracked mHealth data from Clue by BioWink~\cite{clue_app}, one of the most popular and accurate menstrual trackers worldwide~\cite{moglia_evaluation_2016}. To assess our model's predictive ability, we compare it against alternative baselines, including predicting next cycle length based on the mean or median of each user's training cycles, as well as neural network-based baselines. We find that our model's predictive performance is superior, especially as we update predictions over subsequent days of the next cycle. We also generate daily predictions of both cycle length and how likely a user is to skip tracking a cycle, which can aid menstruators in better understanding their cycle timing, allow researchers to identify tracking artifacts, and enable mHealth app developers to enhance app design for improved user engagement.


\section{Results}
\label{sec:results}

We utilize the first $11$ self-tracked cycles from a cohort of $186,108$ Clue users for our analysis of menstruator data. See Table~\ref{tab:data} for summary statistics and Methods for cohort definition details.

\begin{table*}[!h]
	\caption{Summary statistics for selected self-tracked menstruator dataset}
	\label{tab:data}
	\begin{center}
		\begin{tabularx}{\textwidth}{|X|c|X|}
			\hline
			Summary statistic & Selected cohort & Selected cohort (first $11$ cycles only) \\ \hline
			Total number of users & 186,108 & 186,108\\
			Total number of cycles & 3,857,535 & 2,047,166\\
			Number of cycles (mean$\pm$sd, median) & 20.73$\pm$8.35, 18.00 & 11.00$\pm$0.00, 11.00\\
			Cycle length (mean$\pm$sd, median)  & 30.75$\pm$7.73, 29.00 & 30.71$\pm$7.90, 29.00\\
			Period length (mean$\pm$sd, median)  & 4.06$\pm$1.76, 4.00 & 4.13$\pm$1.80, 4.00\\
			Age (mean$\pm$sd, median)  &27.00$\pm$3.74, 27.00 & 27.00$\pm$3.74, 27.00\\
			\hdashline
		\end{tabularx}
	\end{center}
	Summary statistics for selected self-tracked menstruator dataset for the whole dataset, as well as the selected first $11$ cycles only. Average number of cycles, cycle length, period length, and age statistics are per-user. Total number of users and age are the same for the selected cohort and selected cohort's first $11$ cycles only, since they represent the same set of users. We see that cycle length and period length statistics differ very minimally between the selected cohort and the selected cohort's first $11$ cycles only, indicating that using the first $11$ cycles is a reasonable representation of each user's history. Note that the Clue app does not collect information from users on race or ethnicity. 
\end{table*}

We showcase our proposed model's ability to outperform alternative baselines on the menstruator data, especially on later days of the next cycle. This demonstrates the benefit of being able to dynamically update beliefs about both cycle length and likelihood of cycle skips and can help users better understand their cycles as they proceed. In addition, we examine the effect of individual variability on cycle length predictions and the importance of considering individual experiences. We also demonstrate our model's ability to successfully detect self-tracking artifacts on simulated data, which can be utilized in mHealth apps to alert users of possible missed tracking. Finally, we examine the probability distribution that characterizes predicted cycle lengths (the posterior predictive distribution) and showcase how our proposed model lends itself to accurately characterizing the underlying data distribution. 

\paragraph{Proposed model outperforms baselines in cycle length prediction, particularly as cycle proceeds.}
Our generative model allows us to make predictions on each day of the next cycle as it proceeds; we refer to this day as $d_{current}$. On day 0 of the next cycle ($d_{current}=0$), our model outperforms all alternative baselines, as seen in Table~\ref{tab:day_0}.
This performance is consistent across different dataset sizes; for instance, we achieve similar root mean square error, or RMSE (see Methods for mathematical definition), on a smaller subset of individuals, about 20\% of the full dataset size (see Table~\ref{tab:day_0}). For further demonstration of our model's robustness to dataset size (across number of individuals, $I$ and cycles per individual, $C$) and different alternative baselines (i.e., different neural network settings), see Section~\ref{sec:suppl_results} in Appendix. 

\begin{figure}[!h]
	\centering
	\centering
	\includegraphics[width=0.8\linewidth]{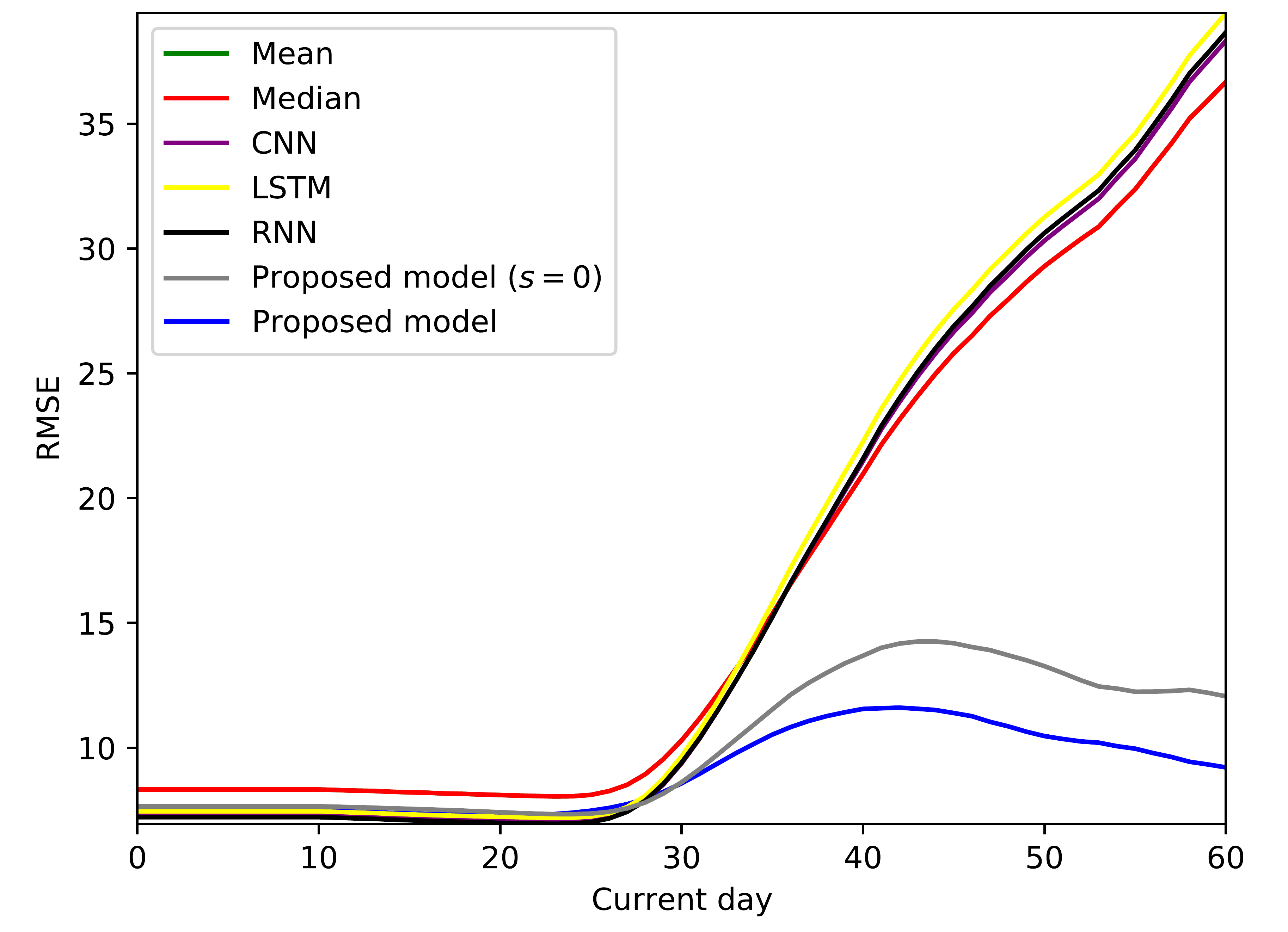}
	\vspace{-1.5ex}
	\caption{Prediction RMSE for proposed model and baselines over current day of the next cycle on the menstruator data. "Proposed model ($s=0$)" indicates an alternative version of our proposed model, assuming the next observed cycle contains no self-tracking artifacts; "Proposed model" indicates the full version of our proposed model, accounting for the presence of potential self-tracking artifacts. Both models' superior performance is magnified past around day 30 of the next cycle; they are able to update predictions dynamically, as compared to static baselines. In particular, accounting for skipped cycles (full version of our proposed model, blue line) proves especially beneficial to prediction accuracy versus assuming the next reported cycle is truth (alternative version of our proposed model, gray line) --- by anticipating the possible presence of skipped cycles, we are able to make more accurate predictions and avoid the bump in RMSE seen in the gray line.}
	\label{fig:pred_by_day}
\end{figure}

Our model's superior performance relative to baselines is especially apparent as the cycle proceeds, as seen in Figure~\ref{fig:pred_by_day} and specific RMSE values on day 40 of the next cycle in Table~\ref{tab:day_40}. In Figure~\ref{fig:pred_by_day}, we see that past $d_{current}=30$, our models (gray line, $s=0$ and blue line, $s\geq0$) display much lower RMSE than baselines. Note that $s$ represents the number of possible skipped cycles in the observed cycle length --- $s=0$ indicates that we assume the next observed cycle length to be the true cycle length, while $s\geq 0$ indicates that we account for the possibility of the observed cycle length containing skipped tracking. In particular, accounting for potential skipped cycles (blue line) proves more advantageous as the cycle proceeds, in comparison to assuming the next observed cycle contains no self-tracking artifacts (gray line). Our model's ability to outperform baselines as the cycle proceeds demonstrates the value in being able to explicitly condition on cycle day and dynamically update predictions, a benefit offered by our proposed generative model.

\begin{table*}[!h]
	\caption{Prediction RMSE results by model on day 0}
	\label{tab:day_0}
	\begin{center}
		\begin{tabularx}{\textwidth}{|X|c|c|}
			\hline
			Model & I = 37K & I = 186K\\ \hline
			Mean & 7.602 & 7.497\\
			Median & 7.586 & 7.489\\
			CNN & 8.102 & 8.027 \\
			LSTM & 7.548 & 7.402\\
			RNN & 7.597 & 7.763\\
			Proposed model (predict with $s=0$) & 7.712 & 7.562\\
			Proposed model & 7.483 & 7.382\\
			\hdashline
		\end{tabularx}
	\end{center}
	Prediction RMSE for proposed model and baselines on day 0 for a subset of the menstruator data, $I=37,222$, and the full menstruator data, $I=186,108$. Note that here we train on $C=10$ cycles and predict the next one. "Proposed model ($s=0$)" indicates an alternative version of our proposed model, assuming the next observed cycle contains no self-tracking artifacts; "Proposed model" indicates the full version of our proposed model, accounting for the presence of potential self-tracking artifacts. Our model outperforms summary statistic-based and neural network-based baselines on day 0 when we account for skipped cycles and does so on only a subset of the data. 
\end{table*}

\begin{table*}[!h]
	\caption{Prediction RMSE results by model on day 40}
	\label{tab:day_40}
	\begin{center}
		\begin{tabularx}{\textwidth}{|X|c|c|}
			\hline
			Model & I = 37K & I = 186K\\ \hline
			Mean & 22.276 & 21.915\\
			Median & 23.675 & 23.394\\
			CNN & 24.741 & 24.506 \\
			LSTM & 23.025 & 22.681\\
			RNN & 23.474 & 22.954\\
			Proposed model (predict with $s=0$) & 15.114 & 14.778\\
			Proposed model & 11.840 & 11.774\\
			\hdashline
		\end{tabularx}
	\end{center}
	Prediction RMSE for proposed model and baselines on day 40 for a subset of the menstruator data, $I=37,222$, and the full menstruator data, $I=186,108$. Note that here train on $C=10$ cycles and predict the next one. "Proposed model ($s=0$)" indicates an alternative version of our proposed model, assuming the next observed cycle contains no self-tracking artifacts; "Proposed model" indicates the full version of our proposed model, accounting for the presence of potential self-tracking artifacts. Our model outperforms summary statistic-based and neural network-based baselines on day 40 when we account for skipped cycles and does so on only a subset of the data. 
\end{table*}

\begin{figure}[!h]
	\centering
	\centering
	\includegraphics[width=0.8\linewidth]{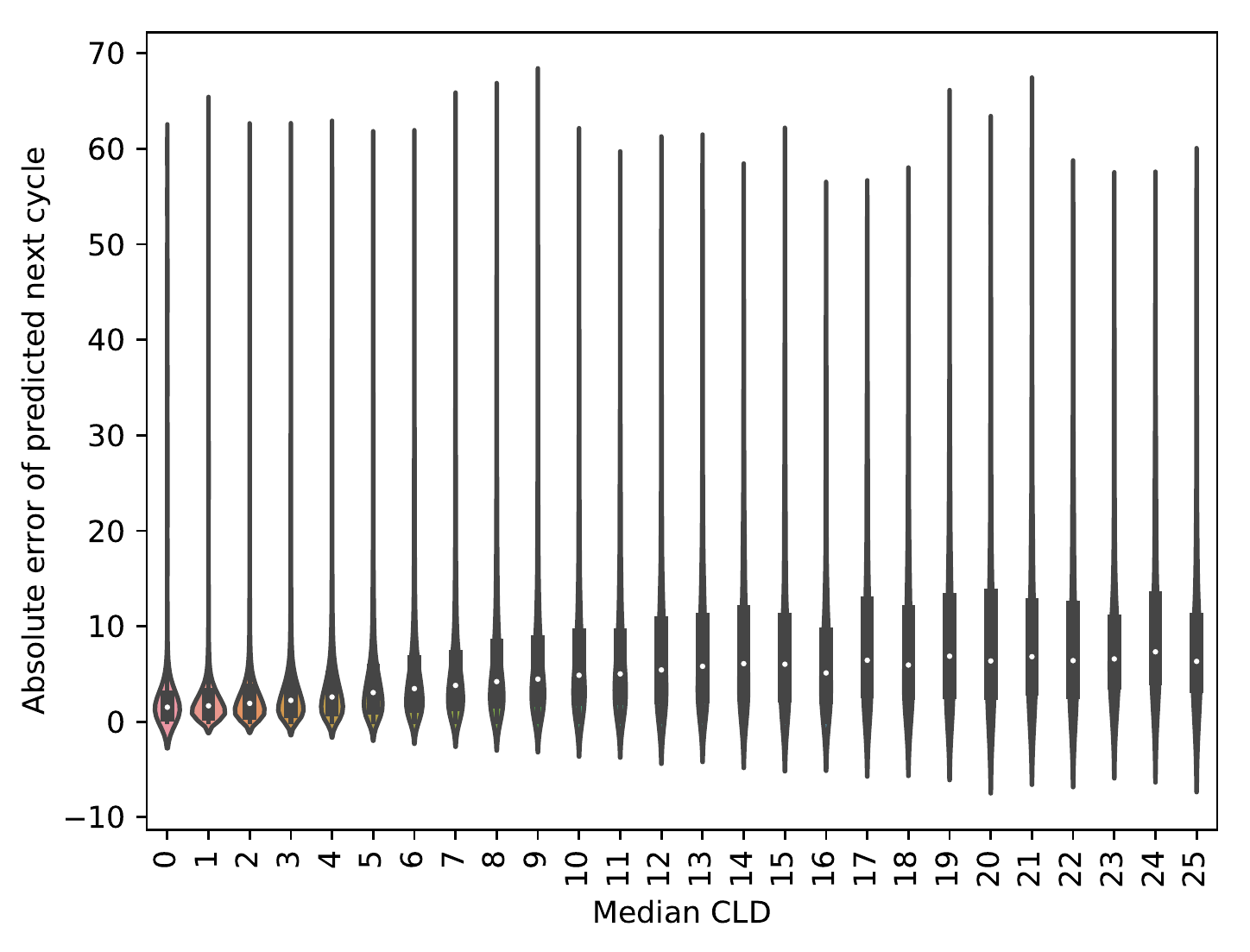}
	\vspace{-1.5ex}
	\caption{Violin plot of absolute error of predicted next cycle length, stratified by user median cycle length difference (CLD) on the menstruator data. We see from the increasing trend in absolute error with median CLD that more variable users are typically more difficult to predict, showcasing that consideration of per-individual behavior is vital to the integrity of our model.}
	\label{fig:reg}
\end{figure}

\paragraph{Variability in cycle tracking history impacts prediction accuracy.}
The menstrual experience is unique, differing within and between individuals. In addition to averaging results over the whole population, we also consider results on an individual level and examine the role that menstrual variability may play in producing accurate predictions. The ability to learn population-wide information while also making individualized predictions is a direct benefit of our hierarchical modeling approach. To assess how predictive accuracy depends on cycle length variability, we showcase a violin plot of per-user median cycle length difference (CLD) versus absolute error in predicted cycle length in Figure~\ref{fig:reg}. Median CLD is a metric for menstrual variability based on previous work~\cite{j-Li2020} --- users with higher median CLD are generally more volatile in their cycle tracking histories, and vice versa (see Methods for mathematical definition). We see that variability impacts prediction accuracy, with more variable users being generally more difficult to predict. This underscores the importance of considering each individual's experience. We also note the presence of outliers within a user's cycle length history, e.g., instances where users may have never skipped in their history, but skip the last cycle. These represent a small proportion of the user base, but skew RMSE computations; for instance, for users with very consistent cycle lengths (i.e., a median CLD of 0), the median absolute error, or MAE (see Methods for mathematical definition) is as low as $1.5$ days, despite the RMSE for this group being $6.15$. 

\begin{figure*}[!h]
	\centering
	\includegraphics[width=0.45\textwidth]{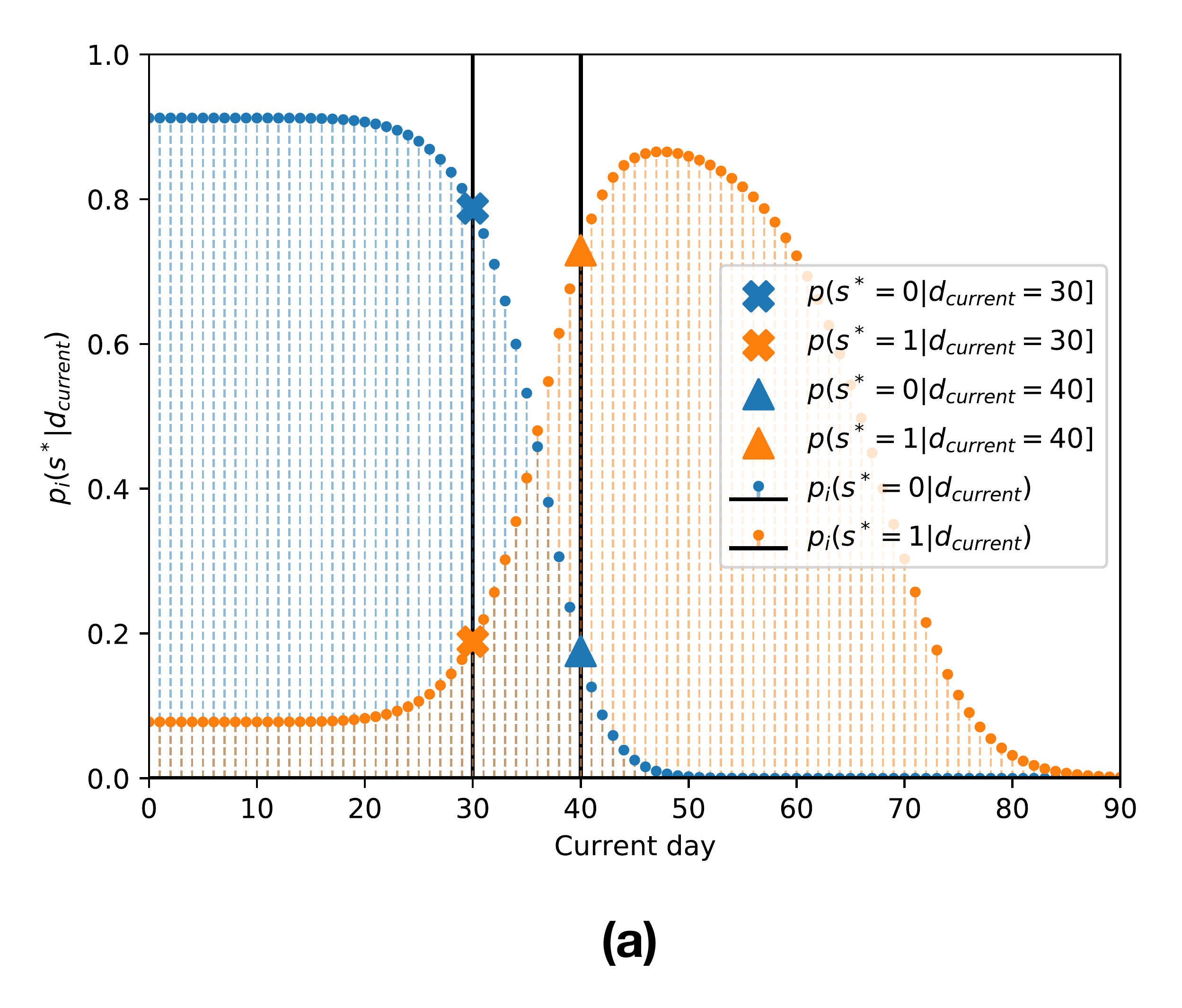}
	\includegraphics[width=0.45\textwidth]{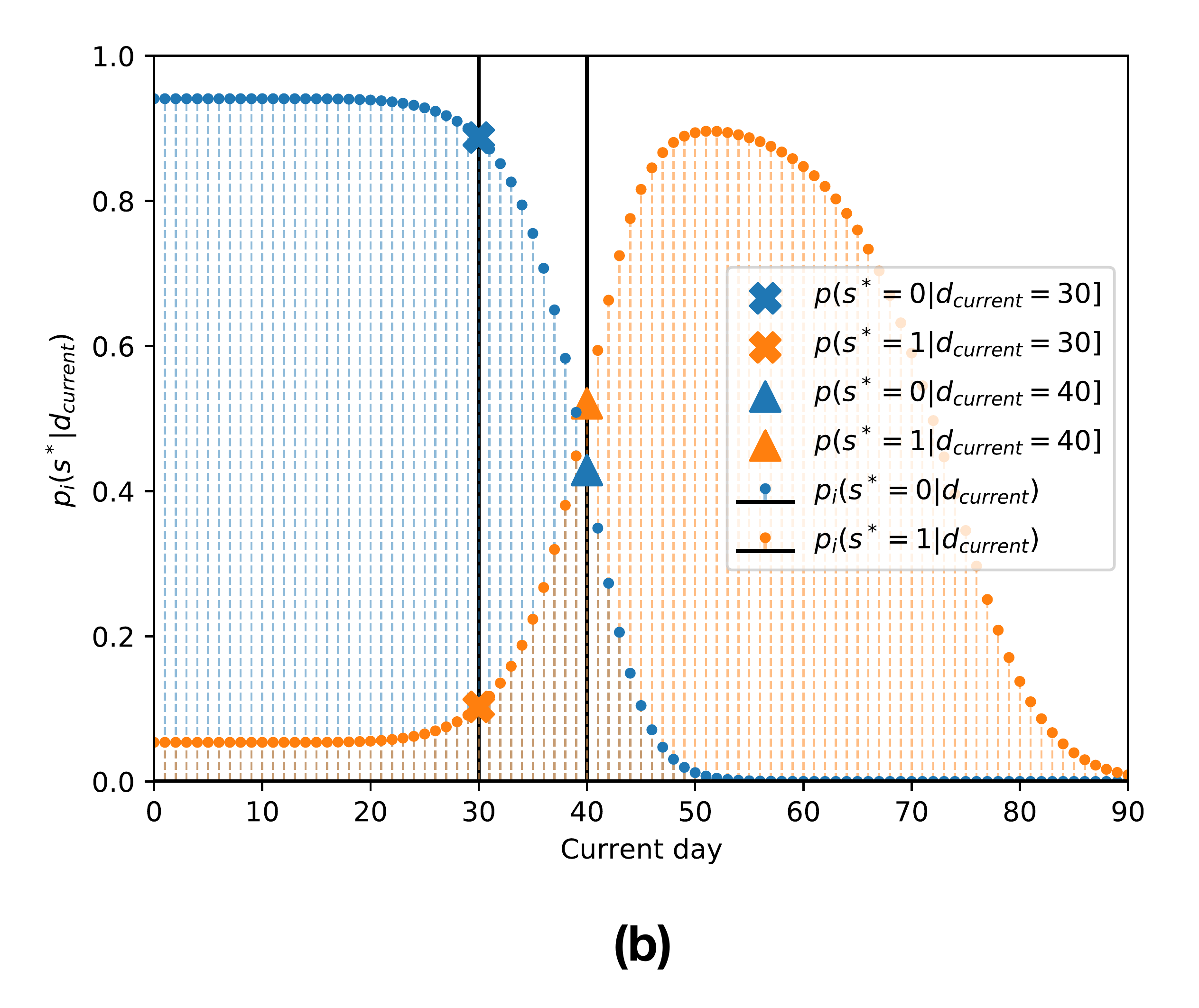}
	\caption{Individual posterior predictive probability of skipping upcoming cycle, $p_i(s^*|d_{current})$, over current day of next cycle $d_{current}$ for two users from simulated data: one who has skipped a cycle in their history (\textbf{a}) and one who has never skipped a cycle (\textbf{b}). Our personalized model detects differences in predicted skipping behavior for the two users. Blue and orange curves represent probabilities of skipping zero or one cycle, respectively; markers indicate probability of skipping zero or one cycle on day 30 or 40 of the upcoming cycle. Note that users can also skip more than one cycle. For both example users, we see that the probability of having skipped zero cycles in the upcoming cycle ($p_i(s^*=0|d_{current})$) is high until day 30. However, past day 30, the model detects that the user \textbf{(a)} who has skipped in their history is more likely to have skipped the upcoming cycle than for the user \textbf{(b)} who has never skipped. This demonstrates how the model takes into account the previous non-skipping behavior of this user. 
	Because data in this experiment is simulated, we know that the user in \textbf{(a)} does actually skip the next cycle, while the user in \textbf{(b)} does not. Our inferred probabilities recover this, showing that our model can accurately detect when a user is likely to have skipped an upcoming cycle based on their individual cycle length histories and update these beliefs over time. 
	}
	\label{fig:skipping}
	\vspace*{-2ex}
\end{figure*}

\paragraph{Accounting for potential cycle skips enables detection of tracking artifacts.}
Identifying skipped cycles is vital to modeling self-tracked cycle lengths accurately, since a skipped period results in an artificially-inflated observed cycle length. Because we define a generative process that explicitly separates cycle patterns from cycle-skipping behavior, we can examine the likelihood of skipped cycles specifically. 
We display our model's ability to detect skipped cycles by illustrating the probabilities of possible cycle skips, shorthanded as $p(s^*|d_{current})$, on simulated data in Figure~\ref{fig:skipping} -- see full modeling and simulated data experiment details in Methods. \textbf{(a)} showcases a simulated user who has skipped in their history, and \textbf{(b)} showcases a simulated user who has never skipped in their history. The vertical lines represent specific days of the next cycle (days 30 and 40), and the markers represent the predicted probability of skipping zero or one cycle on those days.
We choose days 30 and 40 because 30 days is around the average cycle length for these two simulated users, and day 40 represents when the user has surpassed their typical cycle length. These plots showcase how our model is able to detect differences in the underlying skipping phenomena between the two users: as each user passes their `typical' cycle length without tracking, the model adjusts their likelihood of having skipped tracking a cycle based on their previous skipping behavior. Specifically, \textbf{(a)} for the user who has skipped in their history, their probability of skipping one cycle on day 40 
is around 0.8, and their probability of skipping zero cycles on day 40 is around 0.2, a significant drop from a near 0.8 probability on day 30. This showcases how the model is able to incorporate knowledge about this user having previously skipped in computing their propensity to skip their next cycle. In comparison, \textbf{(b)}  the user who has never skipped has a probability of skipping one cycle on day 40 of around 0.5 
--- it is not as clear that this user may have skipped a cycle, because they have never skipped before (\ie this might be an occasional long cycle for this user, which may occur across menstruators in response to other internal or external stimuli). 
While we focus on $s^*=0, 1$ in Figure~\ref{fig:skipping}, note that this behavior holds analogously for $s^*=2$ and beyond. For instance, $p(s^*=2)$ is low early in the next cycle and peaks past day $60$, similar to how $p(s^*=1)$ starts low and peaks past day $30$. The ability to detect and alert users of potential tracking artifacts is important not only to accurately predicting the occurrence of the next cycle, but also to improving the design of mHealth apps as well as the quality of mHealth data for menstrual health research. 
 
\begin{figure*}[!h]
	\centering
	\includegraphics[width=0.45\textwidth]{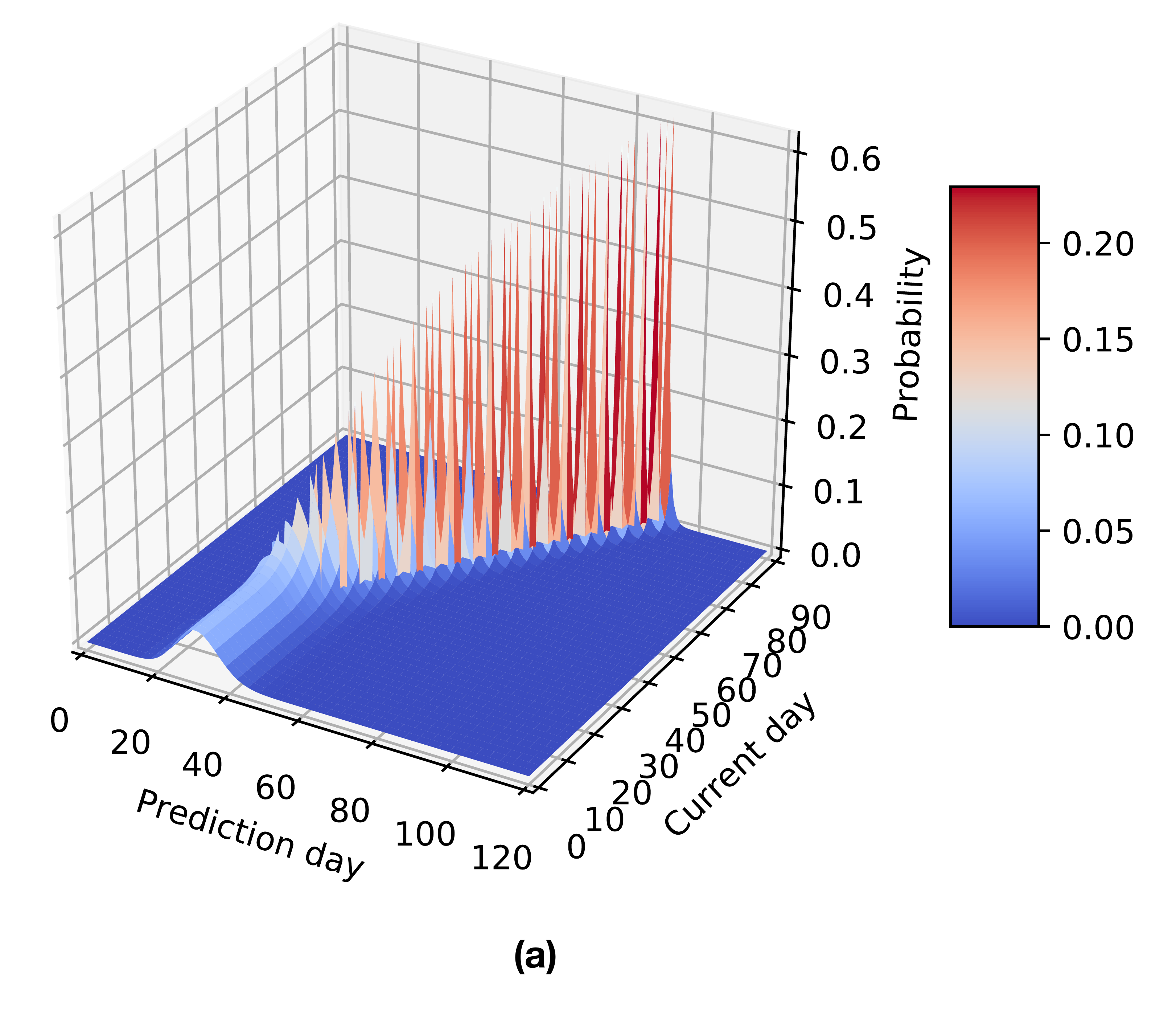}
	\includegraphics[width=0.45\textwidth]{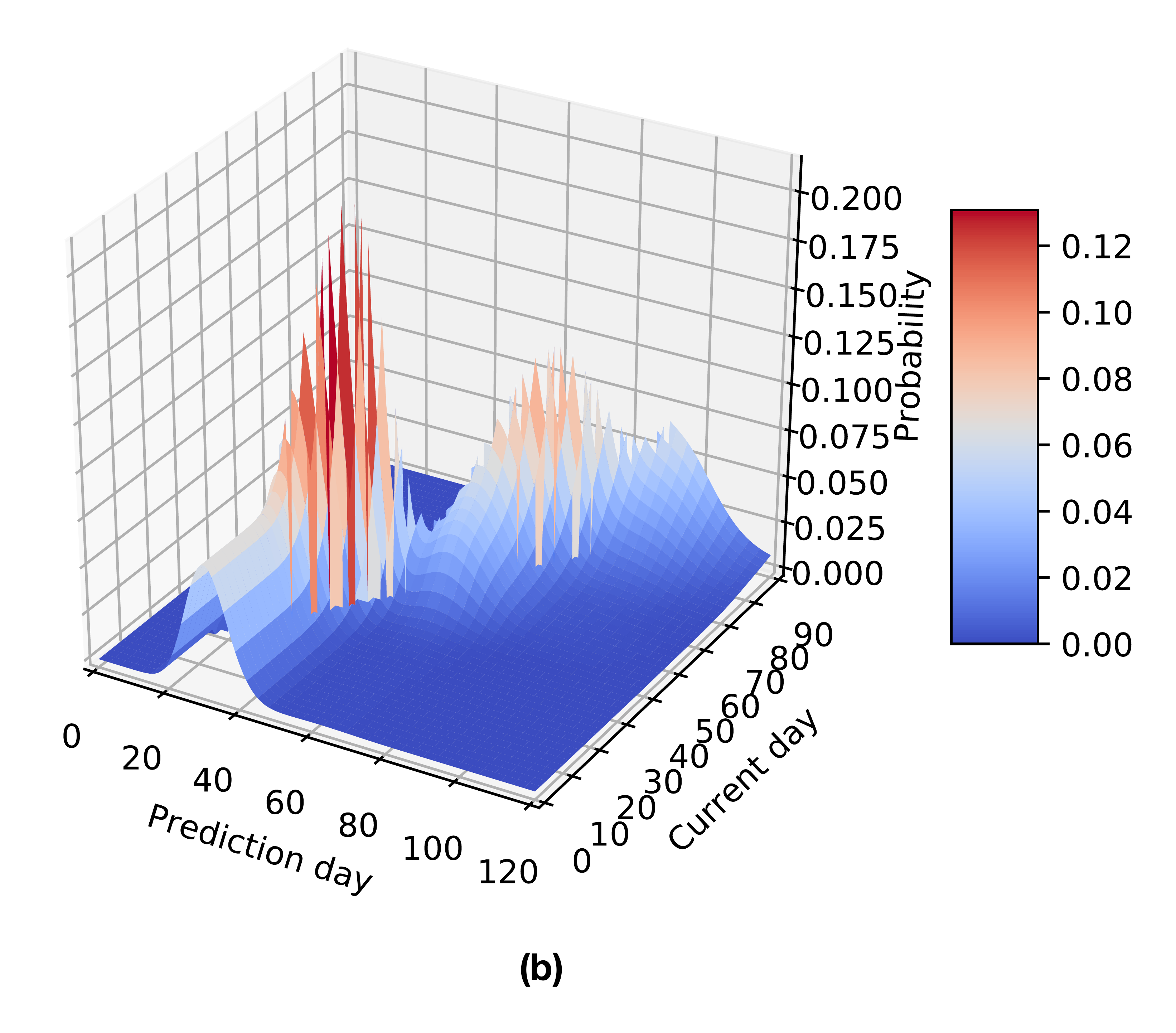}
	\caption{Posterior predictive distribution for cycle length over prediction day $d^*$ (i.e., what the next reported cycle is predicted to be) and current day $d_{current}$ (i.e., day in next cycle) for the same user from menstruator data, assuming either that next observed cycle is truth (\textbf{a}) or that next observed cycle may contain skipped cycles (\textbf{b}).
		\textbf{(a)} When we assume the next observed cycle is true as reported ($s=0$), our posterior predictive distribution is unimodal. The probability of the next cycle length is peaked around 30 until around day 30 of the next cycle, after which the peak moves consistently to the right, indicating that our cycle length predictions are consistently increasing past day 30 and not adjusting for the likelihood of skipped cycles. 
		\textbf{(b)} When we account for the possibility of skipped cycles with $ s\geq 0$, our posterior predictive distribution is multimodal. Prior to day 30 of the next cycle, the distribution is similarly peaked around 30 days, as with the $s=0$ case. However, when the cycle passes day 30, the distribution shows a peak around day 60, indicating the possibility that a user may have skipped a cycle. This behavior holds analogously past day 60. Our explicit modeling of cycle skips allows us to identify when a user may have missed tracking a cycle. 
	}
	\label{fig:posterior}
	\vspace*{-2ex}
\end{figure*}

\paragraph{The posterior predictive distribution for cycle length is interpretable and representative of data.}
In Figure~\ref{fig:posterior}, we showcase our model's posterior predictive distribution for cycle length $p(d^*|\hat u, d_i, d^*>d_{current})$, i.e., the probabilistic next cycle length predictions provided by our model, for a specific user (learned as per their previous cycle length history) as the days of the next cycle proceed.
In particular, Figure~\ref{fig:posterior} shows the probability (z-axis) of a user's next cycle being of an specific length (x-axis) for the current day of the cycle (y-axis), assuming \textbf{(a)} that their next observed cycle is truth (no skipped cycles, $s=0$) or \textbf{(b)} that their next observed cycle may contain skipped cycles (possible skipped cycles, $s\geq 0$).

We are able to accurately update our model's cycle length predictions by updating its beliefs about the likelihood of skipping a cycle over time.
When \textbf{(a)} we assume the next cycle is truth, the posterior predictive distribution is unimodal; however, when \textbf{(b)} we assume the next cycle may not be truth, the posterior predictive distribution is multimodal, with peaks around $d^*=30, 60, 90$.
Such multimodality occurs as a result of (i) conditioning on the day of the next cycle $d_{current}$ and (ii) the explicit modeling of cycle skips, $s$. 
This multimodal distribution mirrors the skipping phenomena observed in the dataset --- when a user passes their `typical' cycle length (around 30 days), they may have skipped tracking of a cycle. The multimodal posterior predictive distribution is not only easily interpretable, but is also crucial to representing self-tracking artifacts in mHealth data and providing accurate cycle length predictions.

\section{Discussion}
\label{sec:discussion}


A hierarchical, generative model offers opportunities to characterize the underlying mechanisms of the varied experience of menstruation as collected via mHealth apps, the first step to a deeper understanding of menstruation as a whole. 
By developing a generative model, we are able to interpret the variables learned from per-user cycle length histories --- we can trace a prediction of cycle length back to the hyperparameters, parameters, and latent variables that underlie it.

Other attempts to model menstrual cycle lengths using user-reported data focus on issues like how to represent between-women and within-women variability. Researchers have represented this variability utilizing hierarchical models~\cite{10.1093/biostatistics/kxq020}, as well as mixture models of standard cycles (cycles 43 days and shorter) and nonstandard cycles (cycles longer than 43 days) ~\cite{10.1093/biostatistics/kxi043}. While these studies capture many important aspects of menstruation (such as the importance of considering each woman's individual cycle behavior) and include exclusion criteria for women who they suspect may not have reported their cycles accurately, they do not explicitly address the user adherence issues encountered when using self-reported mHealth data. Without this consideration, it may be difficult to determine whether nonstandard cycles are the result of individuals skipping tracking. In addition, the definition of a standard or nonstandard cycle may be limiting in itself, and these studies may also be limited in size or scope of the dataset used. For instance, one advantage of our analysis is that we are able to utilize a large dataset of natural menstrual cycles only.

Using observed cycle lengths as our only data source allows us to achieve comparable error to prior studies. In~\cite{time-series} for instance, an RMSE of 1.6 is achieved; however, this RMSE is based on standard cycles only and uses self-tracking data from a mHealth app designed for female athletes, a specific subset of individuals that does not necessarily represent the diversity of women. In our study, when we consider non-variable cycles only (based on the definition of menstrual regularity as represented in Figure~\ref{fig:reg}), our model is able to achieve a similar median absolute error of 1.5 days, but the presence of outliers in more broadly used mHealth apps like Clue (due to unexpected cycle skips) increases the RMSE. 



The hierarchical nature of our model means that once we infer population-level hyperparameters, we can make individualized predictions for new users without using other users' cycle length information. 
Furthermore, we have seen that our model is robust to dataset size --- performance is consistent across relatively small sample sizes (see Appendix Section~\ref{sec:suppl_results} for details). 
This is advantageous because once we have reasonable population-level hyperparameter estimates, there is no need to retrain our model. Therefore, our hierarchical model helps account for privacy concerns; as new users join the app, there is no need to retrain our model or share data between users before we can make predictions. This not only accounts for user concerns about how their data is used, but also makes our model more time-efficient --- only a small subset of users is required for training to then make predictions on any number of individuals. 

Our work showcases the importance of considering the potential unreliability of real-world mHealth data and outlines the performance and interpretability benefits of doing so.  Computing both cycle length and cycle skip predictions allows us to confront the nature of real-world mHealth data and has practical applications for mHealth apps. By explicitly accounting for potential cycle tracking artifacts, we can provide estimates of both next cycle length and probability of skipping tracking. This holds many beneficial implications: it aids users in better understanding their own data and menstrual patterns, helps researchers better disentangle true behavior from self-tracking artifacts by lending insight into the underlying structure of the observed data, and informs mHealth app designers about how to improve user adherence. 





Our model for predicting menstrual patterns showcases the potential that self-tracking mHealth data holds to further understanding of previously enigmatic physiological processes. By utilizing a generative model, we have gained insight into the mechanisms of self-tracking behavior, and in particular, users' propensity to skip tracking. Our dual predictions (i.e., predictions of both cycle length and possible cycle skips) offer a practical application for mHealth apps --- rather than providing an option for users to exclude self-identified faulty cycles after the fact, they can proactively alert users when their probability of skipping tracking is high. For instance, users could be alerted when their cycle skipping probability is near a peak, as in Figure~\ref{fig:skipping}. Note that cycle variability is common, and therefore longer cycle lengths can also be the result of physiological phenomena and not just skipped tracking --- this context, captured by our proposed model, can be provided to users in such alerts. This type of informed alerting helps avoid user notification fatigue (i.e., targeting alerts instead of alerting users everyday to ensure they are tracking) and increases efficacy and accuracy of self-reporting, which is crucial to creating more reliable datasets for the future. This is a vital demonstration of the importance of considering the specific nature of mHealth data that not only enables researchers and users alike to better understand menstruation and the underlying reason behind the observed cycle length, but also provides insight for mHealth app developers of how to alert users to possible inconsistent adherence in an informed way. As mHealth apps continue to grow in popularity and serve as an increasingly important source of information for healthcare interventions, it is vital to consider how to improve the quality of mHealth data and ensure it is being treated responsibly; these insights can aid researchers using such data sources to do so. 

There is also great potential to extend our model to include other covariates, namely symptom observations. Including information beyond cycle lengths is crucial to understanding cycle variability more holistically~\cite{vitzthum} and may have significant impact on cycle prediction accuracy. In our previous work~\cite{j-Li2020}, we found that there is a relationship between cycle timing and symptom experiences; other studies have also included symptom covariates, like cramps and period flow, to examine how these impact reported menstrual cycle length~\cite{time-series}. By extending our fully generative model to a hybrid deep generative model with symptom information, we can better understand this relationship from a mechanistic perspective and incorporate more context of each individual's menstrual experience. In particular, the model we've proposed here is the necessary first step to a proven hybrid deep learning approach~\cite{tansey2020doseresponse} that models experimental details in a generative framework and complex biological details in a deep way. While our model is able to outperform baselines by using only previous cycle lengths as predictors for next cycle length, a hybrid model where cycles are generated from the proposed generative model with latent variables and parameters learned by a deep model that uses both symptom (e.g., headache, abdominal pain) and cycle length information as input will allow us to leverage the full potential of large self-tracking datasets to gain further insight into women's menstrual experiences and empower users to better track and predict their cycles. 


\section{Methods}
\label{sec:methods}
\subsection*{Proposed hierarchical, generative model.}
Our hierarchical, generative model for self-tracked menstrual cycle lengths incorporates population-wide knowledge (via informative priors for hyperparameters) and learns individualized cycle length patterns (via per-user parameters and predictions). 

Its hierarchical nature allows for the representation of different levels of information: individual-specific patterns (i.e., user-specific tracking histories), as well as population-wide characteristics (i.e., common patterns that exist across the studied population).  We learn population-level information as population-level distributions and their hyperparameters and characterize the unique nature of each individual's menstrual experience through individual-level parameters and predictions.

\begin{figure}[h!]
	\centering
	\centering
	\includegraphics[width=0.7\linewidth]{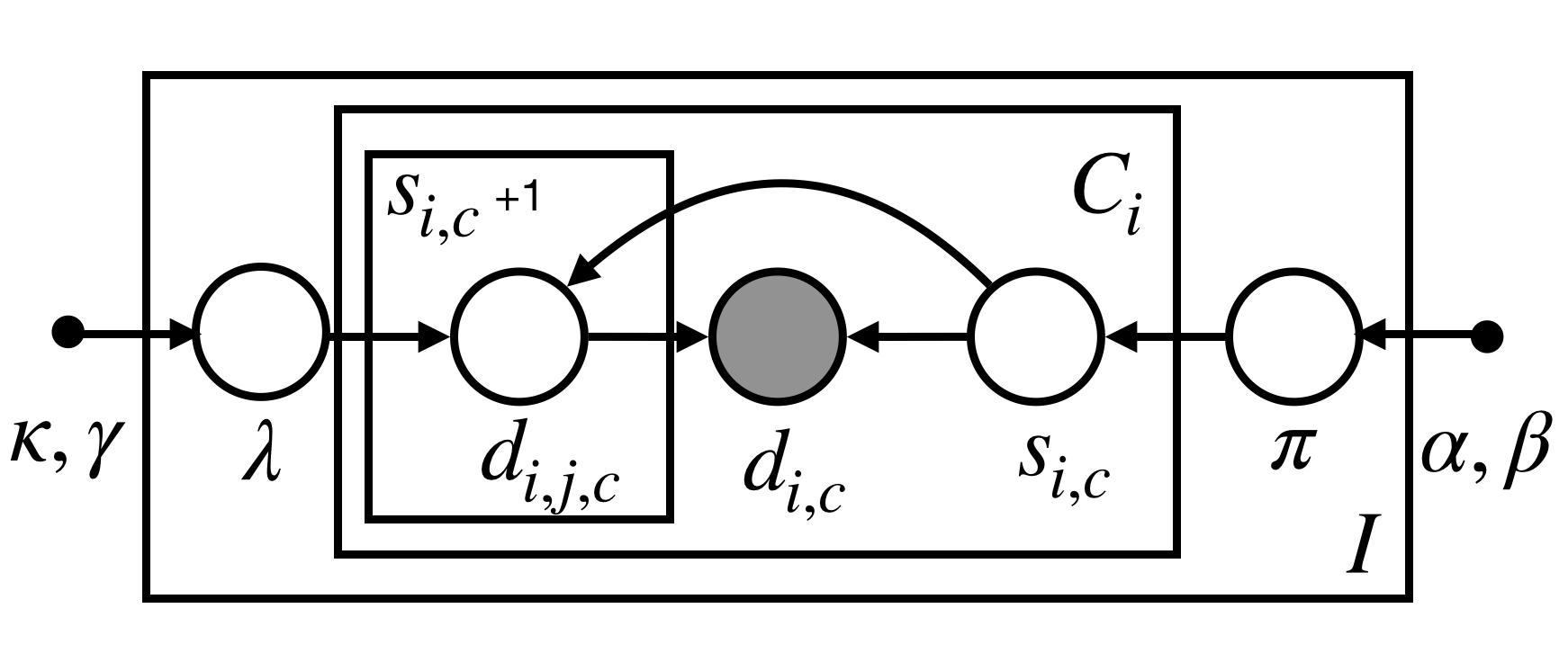}
	\vspace{-1.5ex}
	\caption{Hierarchical graphical model for proposed generative process. Individual-level parameters $\lambda_i$ (average cycle length without skipping) and $\pi_i$ (probability of skipping a cycle) are drawn from population-level distributions characterized by hyperparameters $u = [\kappa, \gamma, \alpha, \beta]$. Shaded circles represent observed data, open circles represent latent variables, and dots represent hyperparameters.}
	\label{fig:graphical_model}
\end{figure}

A generative model indicates that we propose the distributions from which each variable is generated, allowing for greater interpretability. 
The proposed generative model represents expected cycle length patterns $\lambda_i$ for individual $i$ and probability of skipping tracking $\pi_i$ separately, as shown in Figure~\ref{fig:graphical_model}. By doing so, we are able to disentangle true, per-user cycle behavior from self-tracking artifacts; in addition to predicting cycle length,
we can also gain insight into cycle skipping behavior, i.e., our model is interpretable and produces accurate individualized predictions. 

The proposed model outlined in Figure \ref{fig:graphical_model} extends Poisson regression (which represents expected user cycle lengths via parameter $\lambda_i$) with individual latent variables $\pi_i$ that accommodate each user's propensity for skipping tracking.
We decouple true from observed cycle lengths $d_{i,c}\geq0$: the variable $s_{i,c}\geq 0$ indicates the number of not reported (skipped) cycles for user $i$ during observed cycle $c$. We represent each individual's cycle tracking history as $d_i=\{d_{i,c}\} \forall c$ and each user's history of number of skipped cycles as $s_i=\{s_{i,c}\} \forall c$.
The proposed generative process is detailed in Section~\ref{sec:suppl_model} of the Appendix. 

\subsection*{Self-tracked menstruator data.}


We leverage a de-identified self-tracked dataset from Clue by BioWink~\cite{clue_app}, comprised of $117,014,597$ self-tracking events over $378,694$ users. For this full dataset, users have a median age of $25$ years, a median of $11$ cycles tracked, and a median cycle length of $29$ days. Clue app users input overall personal information at sign-up, such as age and hormonal birth control (HBC) type. The dataset contains information from 2015-2018 for users worldwide, covering countries within North and South America, Europe, Asia and Africa. Users can self-track symptoms over time --- for this paper, we focus on period data (i.e., the users' self-reports on which days they have their period). A period consists of sequential days of bleeding (greater than spotting and within ten days after the first greater than spotting bleeding event) unbroken by no more than one day on which only spotting or no bleeding occurred. Note that Clue considers a menses duration longer than 10 days as an outlier, as it would exceed mean period length plus 3 standard deviations for any studied population~\cite{j-Vitzthum2009}.
We only use cycle lengths as input to our proposed model, where we define a menstrual cycle as the span of days from the first day of a period through to and including the day before the first day of the next period ~\cite{j-Vitzthum2009}.  In addition, a user has the opportunity to specify whether a cycle should be excluded from their Clue history --- for instance, if the user feels that the cycle is not representative of their typical menstrual behavior due to a medical procedure or changes in birth control, they may elect to exclude it.

Our cohort consists of users aged 21-33 (because cycles are more likely to be ovulatory and less variable in their lengths during this age interval~\cite{j-Treloar1967,j-Chiazze1968,j-Ferrell2005,j-Vitzthum2009,j-Harlow2012}) with natural menstrual cycles (i.e., no hormonal birth control or intrauterine device). To rule out cases that indicate insufficient engagement with the app, we remove users from our cohort who have only tracked two cycles and cycles for which the user has not provided period data within 90 days. We use the first $11$ cycles for all $186,108$ menstruators with more than $11$ cycles tracked (since $11$ is the median number of cycles tracked in the full Clue dataset). For summary statistics on this selected cohort, see Table~\ref{tab:data}. Results reflected in Figures~\ref{fig:pred_by_day}, ~\ref{fig:reg}, and ~\ref{fig:posterior} and Tables~\ref{tab:day_0} and ~\ref{tab:day_40} use this real menstruator dataset.

\subsection*{Simulated data.}
In order to assess the ability of our model to recover skipped cycles, we separately train our model on simulated cycle length data for $10,000$ users (with $C=10$ cycles each), generated from our proposed generative process. We then take two cohorts of users: those who have never skipped a cycle in their history, and those who have skipped a cycle in their history. Note that we have access to ground truth cycle length and skipping information in this simulated case. For a sample user from each of these cohorts, we predict their probabilities of possible cycle skips $p(s^*|\hat u, d_i, d^*>d_{current})$ for the $11$th cycle, utilizing the inferred population-wide hyperparameters $\hat u$ and individual cycle length histories $d_i$. Results in Figure~\ref{fig:skipping} use this simulated dataset.

\subsection*{Model training.}
We train our model on the full eligible dataset of $186,108$ users described in Table~\ref{tab:data}, training on the first $10$ cycles and predicting the quantities of interest for the $11$th cycle of each user.
Population-level hyperparameter inference is achieved via type-II maximum likelihood estimation. Expert knowledge is incorporated via hyperparameter priors. The data log-likelihood under the proposed model is computed by a Monte Carlo approach (see Section~\ref{sec:suppl_model} in Appendix for details). 



We have evaluated the robustness of our training and predictive performance with respect to the dataset size (both with respect to individuals in the population, as well as cycles per individual) in order to quantify the predictive power of different modeling choices. We find that our model is generally stable across different training set sizes (see Section~\ref{sec:suppl_results} in Appendix for details).
To account for possible time dependency of tracked cycles, we also experimented with shuffling the order of each user's cycles and found no significant difference in results, i.e., the predictive performance of the model is not dependent on the specific cycle length ordering (see Section~\ref{sec:suppl_results} in Appendix for details).

\subsection*{Individualized predictions.}
Our generative model allows for computation of both each user's cycle length and cycle skip posterior predictive distributions, which are updated online. For next cycle length $d^*$ prediction, we compute $p(d^*|\hat u, d_i, d^*>d_{current})$, where $\hat u$ represents learned hyperparameters, $d_i$ refers to the individual's cycle length history, and $d_{current}$ is the current day of the next cycle (i.e., if $d_{current}=1$, then $d^*$ must be at least 2). Our proposed generative model allows for updating predictions online by conditioning on the current day of the next cycle (see posterior predictive distribution in Figure~\ref{fig:posterior}). For pointwise predictions updated by day ($d_{current}$), we use the expected value $E[d^*|\hat u, d_i, d^*>d_{current}]$. Beyond cycle length information, our model allows for computation of the probability of each user skipping tracking of the next cycle: $p(s^*|\hat u, d_i, d^*>d_{current})$, also conditioning on learned hyperparameters, user's cycle length history only, and current day of the cycle.

To compute cycle length predictions, our model offers two possibilities:
1) set $s=0$ and assume that the next reported cycle will be truth (i.e., that the next observed cycle will not be skipped); or
2) set $s \geq 0$ and integrate it out, assuming the next reported cycle may not be truth (i.e., accounting for the user possibly skipping their next cycle tracking).
We evaluate and show the performance of our model for these two versions, with details on the prediction procedure provided in Section~\ref{sec:suppl_model} of the Appendix .


\subsection*{Evaluation metrics.}

We use root mean square error (RMSE) to evaluate the average prediction accuracy of our model across all users. RMSE of $N$ actual data points $d_i$ and predictions $\hat d_i$ is computed as

\begin{equation}
	\text{RMSE} = \sqrt{\frac{\sum_{i=1}^N\left(d_i - \hat d_i\right)^2)}{N}}
\end{equation}

We use absolute error and median absolute error (MAE) to evaluate prediction accuracy of our model on a per-user basis, as in Figure~\ref{fig:reg}. Absolute error between an actual data point $d_i$ and prediction $\hat d_i$ is computed as

\begin{equation}
	\text{Absolute error} = |d_i - \hat d_i|
\end{equation}

We use median CLD as a metric for evaluating menstrual regularity, based on previous work characterizing menstruation~\cite{j-Li2020}. CLDs are computed per-user as the absolute differences between consecutive cycle lengths -- for a user whose $C$ cycles are defined as $d = [d_0, d_1, d_2, ... d_C]$, the CLDs are computed as

\begin{equation}
	\text{CLDs} = [|d_1 - d_0|, |d_2 - d_1|, ... |d_C - d_{C-1}|]
\end{equation}

For instance, a user with cycle lengths $d=[30, 40, 25, 30]$ will have CLDs of $[10, 15, 5]$ and a median CLD of $10$.

\subsection*{Alternative baselines.}
To evaluate the predictive performance of our proposed model, we consider summary statistic-based and neural network-based baselines (all details are provided in Section~\ref{sec:suppl_results} of the Appendix):
\begin{itemize}
	\item Mean and median baselines: the predicted next cycle for each user is the average (or median) of their previously observed cycle lengths.
	\item CNN: a 1-layer convolutional neural network with a 3-dimensional kernel.
	\item RNN: a 1-layer bidirectional recurrent neural network with a 3-dimensional hidden state.
	\item LSTM: a 1-layer Long Short-Term Memory neural network with a 3-dimensional hidden state.
\end{itemize}

Note that we also test other neural network architectures (increasing number of layers and changing kernel or hidden state dimensionality) and find no meaningful performance difference --- see Section~\ref{sec:suppl_results} of the Appendix. 





\section{Data availability}
\label{sec:data availability}
The database that supports the findings of this study was made available by Clue by BioWink.
While it is de-identified, it cannot be made directly available to the reader.
Researchers interested in gaining access to the data should contact Clue by BioWink to establish a data use agreement.  

\section{Acknowledgements}
The authors are deeply grateful to all Clue users whose de-identified data have been used for this study.


\section{Author contributions}
KL, IU, CHW, and NE conceived the proposed research and designed the experiments. KL and IU processed the dataset and conducted the experiments. KL wrote the first draft of the manuscript. IU, CHW, AD, AS, VJV, and NE reviewed and edited it. All authors read and approved the manuscript.

\section{Competing interests}
KL is supported by NSF's Graduate Research Fellowship Program Award \#1644869.
IU and NE are supported by NLM award R01 LM013043.
KL, IU, CW, and NE declare that they have no competing interests. AS and VJV were employed by Clue by BioWink at the time of this research project.

\bibliographystyle{naturemag}

\clearpage
\appendix
\section{The generative model}
\label{sec:suppl_model}

We provide details on the generative process for self-tracked mHealth cycle lengths, which draws per-user specific parameters from population level shared priors:

\begin{itemize}
	\item \textbf{Observed variables:} Observed cycle length $d_{i,c}$, with $c=\{1,\cdots,C_i\}$ cycle lengths for each individual $i=\{1,\cdots, I\}$.
	Each true cycle length (for user $i$, cycle $c$, out of the number of skipped cycles $j$) is drawn from a Poisson distribution, $d_{i,j,c} \sim p(d_{i,j,c}|\lambda_i) = Pois(d_{i,j,c}|\lambda_i)$.
	The sum of independent Poissons is a different Poisson distribution, so the observed cycle length ($d_{i,c} = \sum_{j=0}^{s_{i,c}+1} d_{i,j,c}$) is also drawn from a Poisson, conditioned on the number of skipped cycles,
	\begin{equation}
		d_{i,c}\sim Pois(\lambda_i(s_{i,c}+1)) \;.
	\end{equation}
	\item \textbf{Latent variables:} $s_{i,c}$ denotes the number of skipped (not reported) cycles, with $c=\{1,\cdots, C_i\}$ cycle lengths for each individual $i=\{1,\cdots, I\}$.
	The number of skipped cycles is drawn from a truncated Geometric distribution with a maximum number of skipped cycles $S$, 
	\begin{equation}
		s_{i,c} \sim p(s|\pi_i)=\frac{\pi_i^s (1-\pi_i)}{\sum_{s=0}^{S}\pi_i^s (1-\pi_i)} =\frac{\pi_i^s }{\sum_{s=0}^{S}\pi_i^s } =\frac{\pi_i^s (1-\pi_i)}{(1-\pi_i^{(S+1)})} \; \text{ for } s \in \mathbb{N} \;.
		\label{eq:probability_of_skipping}
	\end{equation}
	\item \textbf{Parameters $\lambda_{i}$:} the Poisson rate parameters for each individual $i=\{1,\cdots, I\}$.
	Per-user Poisson rate parameters $\lambda_i$ are drawn from a population-level Gamma distribution
	\begin{equation}
		\lambda_i \sim p(\lambda|\kappa,\gamma)=\frac{\gamma^\kappa}{\Gamma(\kappa)}\lambda^{\kappa-1}e^{-\gamma \lambda}
		\quad \text{ for } \lambda > 0 \text{ and } \kappa, \gamma > 0.
	\end{equation}
	
	\item \textbf{Hyperparameters of the Poisson rate parameter:} $\kappa$, $\gamma$ of a Gamma distribution prior for the Poisson rate at the population level.
	\item \textbf{Parameters $\pi_{i}$:} the probability of skipping a cycle for each individual $i=\{1,\cdots, I\}$.
	The probability of an individual skipping a cycle is drawn from a population-level Beta distribution
	\begin{equation}
		\pi_i \sim p(\pi|\alpha,\beta)=\frac {\Gamma(\alpha+\beta)}{\Gamma(\alpha)\Gamma(\beta)} \pi^{\alpha-1}(1-\pi)^{\beta-1} \;, 
		\quad \text{ for } \pi \in [0,1] \text{ and } \alpha, \beta > 0.
	\end{equation}
	\item \textbf{Hyperparameters of the geometric distribution parameters:} $\alpha$, $\beta$ of the population level Beta distribution prior on skipping probabilities.
\end{itemize}

\subsection*{Inference details}
\label{sec:suppl_inference}
Given a dataset of $C_i$ cycle lengths for $I$ users, we perform hyperparameter inference via type-II maximum likelihood estimation.
We compute a Monte Carlo (MC) approximation to the negative log-likelihood: $-\ln(p(d|u)) = -\ln(\sum_i p(d_i|u))$. Due to the impossibility of integrating out the number of skipped cycles $s_{i,c}$ analytically, we compute a MC approximation to each cycle length likelihood $p(d_i|u)$ with $M$ samples,
\begin{equation}
	p(d_i|u) = \frac{1}{M}\sum_m p(d_i|\theta_m) \;, \theta_m \sim p(u)
\end{equation}

where $u$ represents the hyperparameters $[\alpha, \beta, \kappa, \gamma]$ of the distributions from where samples $\theta_m$, representing the parameters $[\lambda_m, \pi_m]$, are drawn. We compute the probability $p(d_i|\theta_m)$ by integrating out the probability of skipping $s_{i,c}$, which is drawn from a truncated geometric distribution as in Eqn.~\eqref{eq:probability_of_skipping}:

\begin{subequations}
	\begin{align}
		p(d_i|\theta_m) &= \prod_{c=1}^{C_i} p(d_{i,c}|\theta_m) = \prod_{c=1}^{C_i} \sum_{s=0}^S p(d_{i,c}|\lambda_m, s)p(s|\pi_m)\\
		&= \prod_{c=1}^{C_i} \sum_{s=0}^S ((\lambda_m(s+1))^{d_{i,c}}e^{-\lambda_m(s+1)}/d_{i,c}!)\left(\frac{\pi_m^s(1-\pi_m)}{\sum_{s=0}^S \pi_m^s(1-\pi_m)}\right)\\
		&= \prod_{c=1}^{C_i} \frac{\lambda_m^{d_{i,c}}e^{-\lambda_m}}{d_{i,c}!} \sum_{s=0}^S ((s+1)^{d_{i,c}}e^{-\lambda_m s})\left(\frac{\pi_m^s}{\sum_{s=0}^S \pi_m^s}\right)\\
		&= \prod_{c=1}^{C_i} \phi(\lambda_m) \frac{\sum_{s=0}^S (s+1)^{d_{i,c}}(\pi_m e^{-\lambda_m})^s}{\sum_{s=0}^S \pi_m^s}\\
		&= \prod_{c=1}^{C_i} \phi(\lambda_m) \frac{\sum_{s=0}^S (s+1)^{d_{i,c}}(\pi_m e^{-\lambda_m})^s}{\frac{1-\pi_m^{S+1}}{1-\pi_m}}\\
		&= \prod_{c=1}^{C_i} \frac{1-\pi_m}{1-\pi_m^{S+1}} \phi(\lambda_m) \sum_{s=0}^S (s+1)^{d_{i,c}}(\pi_m e^{-\lambda_m})^s
	\end{align}
	\label{eq:marginalized_user_likelihood}
\end{subequations}

where $d_{i,c}$ represents one cycle length $c$ for a given user $i$, $C_i$ is the number of cycles for user $i$, $S$ is the maximum value of $s$, and $\phi$ is the Poisson density. 

\subsection*{Prediction details}
\label{sec:suppl_prediction}
In order to update our predictions of per-user cycle length as each subsequent day passes, we are interested in the posterior of the next reported cycle length $d^*$, conditioned on previous cycle lengths $d_i$ for a user $i$ and the day of the current cycle $d_{current}$ ,

\begin{subequations}
	\begin{align}
		p(d^*|d^*>d_{current}, d_i, \hat u) &= \frac{p(d^*, d^*>d_{current}|d_i, \hat u)}{p(d^*>d_{current}|d_i, \hat u)} = \frac{p(d^*|d_i, \hat u)I(d^*>d_{current})}{p(d^*>d_{current}|d_i, \hat u)}
	\end{align}
\end{subequations}
where we explicitly indicate that $p(d^*, d^*>d_{current}|d_i, \hat u) = 0$ if $d^*\leq d_{current}$.

In addition to characterizing the full distribution, we are interested in computing the expectation of the conditional predictive posterior as a point estimate for the next cycle length,

\begin{subequations}
	\begin{align}
		E[p(d^*|d^*>d_{current}, d_i, \hat u)] &= \sum_{d^*} d^* p(d^*|d^*>d_{current}, d_i, \hat u) \\
		&= \sum_{d^*} d^*  \frac{p(d^*|d_i, \hat u)I(d^*>d_{current})}{p(d^*>d_{current}|d_i, \hat u)} \\
		&= \frac{\sum_{d^*} d^* p(d^*|d_i, \hat u)I(d^*>d_{current})}{p(d^*>d_{current}|d_i, \hat u)} \\
		&= \frac{\sum_{d^* > d_{current}} d^* p(d^*|d_i, \hat u)}{p(d^*>d_{current}|d_i, \hat u)} \\
		&= \frac{\sum_{d^*=d_{current}+1}^D d^* p(d^*|d_i, \hat u)}{\sum_{d^*=d_{current}+1}^D p(d^*|d_i, \hat u)} 
	\end{align}
\end{subequations}

The key term above is $p(d^*|d_i, \hat u)$:
\begin{equation}
	p(d^*|d_i, \hat u) = \frac{\int d\lambda d\pi q(\lambda)b(\pi)\sum_{s*} p(s^*|\pi)p(d^*|s^*, \lambda) p(d_i|\lambda,\pi)}{\int d\lambda d\pi q(\lambda) b(\pi) p(d_i|\lambda,\pi)} \;,
\end{equation}

where $d_i$ are the cycle lengths for a user $i$ and $s_i$ are the number of skipped cycles for a user, and $d^*$, $s^*$ are the next reported cycle length and next number of skipped cycles, respectively. For the truncated geometric distribution on skipping probabilities, we compute the above as
\begin{subequations}
	\begin{align}
		p(d^*|d_i, \hat u) &= \frac{\sum_{m=1}^M \frac{1-\pi_m}{1-\pi_m^{S+1}}\sum_{s^*=0}^S \pi_m^{s^*} p(d^*|s^*, \lambda_m) p(d_i|\lambda_m, \pi_m)}{\sum_{m=1}^M p(d_i|\lambda_m, \pi_m)} \;.
	\end{align}
\end{subequations}

We compute $p(d^*|d_i, \hat u)$ for a range of cycle length days $d^*=\{0, \cdots, D\}$, normalizing appropriately over $d^*$ for each value of $d_{current}$, using $p(d_i|\lambda_m, \pi_m)$ (as specified in Eqn.~\eqref{eq:marginalized_user_likelihood} of the description of inference) and $p(d^*|s^*, \lambda_m) = Pois(\lambda(s^*+1))$ (i.e., the Poisson PMF), where we must also normalize $p(d^*|s^*, \lambda)$ over $d^*=\{0, \cdots, D\}$.

\subsection*{Implementation details}
\label{sec:suppl_implementation}

We optimize the negative log-likelihood $-\ln(p(d|u)) = -\ln(\sum_i p(d_i|u))$ with $p(d_i|u)$ as in Eqn.~\eqref{eq:marginalized_user_likelihood} with respect to hyperparameters $u$ via stochastic gradient descent. Specifically, we utilize Adam~\cite{kingma2017adam}, an adaptive gradient method. All models have been implemented using PyTorch, and trained with minibatches of size $100$. All neural network-based models are trained (with dropout) on the observed cycle lengths for the whole cohort. Predictions are based on each per-user available cycle lengths.

Since we sequentially predict next cycle length, our train-test split is over the number of cycle lengths available, i.e., we train the models on $C$ cycles and predict the $C+1$th cycle, where $C=\{2, \cdots, 10\}$.

For reproducibility, we provide the settings for priors, learning rate, and other details for each of the models below:

\begin{itemize}
	\item \texttt{CNN}: number of layers = $1$, kernel size = $3$, stride = $1$, padding = $0$, dilation = $1$, nonlinearity = $tanh$, dropout = $0.9$, training criterion = MSE, epoch convergence criteria as maximum number of epochs = $1000$, loss convergence criteria $\epsilon_{loss} = 1e-3$, optimizer = Adam, learning rate = $0.01$. 
	\item \texttt{RNN}: number of layers = $1$, hidden size = $3$, nonlinearity = $tanh$, dropout = $0.9$, epoch convergence criteria as maximum number of epochs = $1000$, loss convergence criteria $\epsilon_{loss} = 1e-3$, optimizer = Adam, learning rate = $0.01$. 
	\item \texttt{LSTM}: number of layers = $1$, hidden size = $3$, nonlinearity = $tanh$, dropout = $0.9$, epoch convergence criteria as maximum number of epochs = $1000$, loss convergence criteria $\epsilon_{loss} = 1e-3$, optimizer = Adam, learning rate = $0.01$. 
	\item \texttt{Proposed model}: $u_0 = [\kappa_0 = 180, \gamma_0 = 6, \alpha_0 = 2, \beta_0 = 20]$, $S=100$ (for both inference and prediction), $M=1000$ (for both inference and prediction), epoch convergence criteria as maximum number of epochs = $1000$, loss convergence criteria $\epsilon_{loss} = 1e-3$, optimizer = Adam, learning rate = $0.01$. 
	\item \texttt{Proposed model (s=0)}: same as above, with $S=100$ in inference but $S=0$ for next-cycle length prediction.
\end{itemize}

\clearpage
\section{Supplementary results}
\label{sec:suppl_results}
For the results presented in the main text, we utilize a prior $u_0 = [\kappa_0=180, \gamma_0=6, \alpha_0=2, \beta_0=20]$, from which we draw our initial $\theta = [\lambda, \pi]$. This is informed by expert knowledge about average cycle length (around 30 days) and the likelihood of skipping (relatively low) in our dataset.

\subsection*{Performance stability across different priors}
In order to assess the impact of the prior, we evaluate the impact of training the model on different ones, namely a uniform prior on $\pi$ (no prior knowledge on skipping likelihood), as well as a less informative (i.e., flatter) prior on both $\lambda$ and $\pi$.

\begin{figure}[h!]
	\centering
	\centering
	\includegraphics[width=0.4\linewidth]{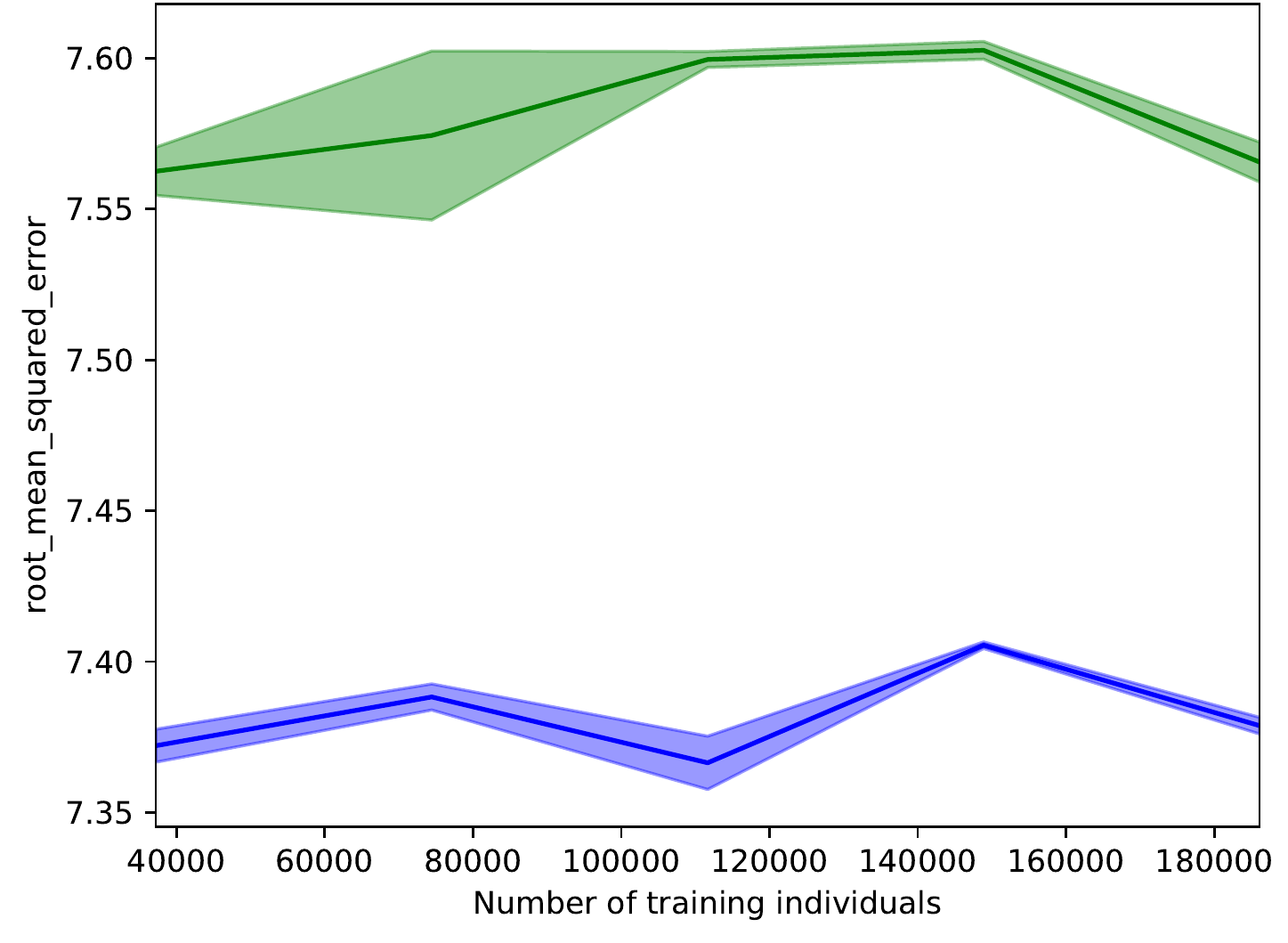}
	\includegraphics[width=0.2\textwidth, trim=0 -19.5em 0 0]{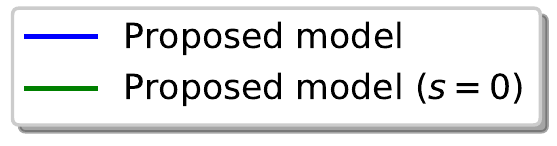}
	\vspace{-1.5ex}
	\caption{Prediction RMSE over number of training individuals for a less informative (i.e., a more uncertain) prior on $\lambda$ and $\pi$, $u_0 = [60, 2, 0.01, 0.1]$.}
	\label{fig:prior_1}
\end{figure}

\begin{figure}[h!]
	\centering
	\centering
	\includegraphics[width=0.4\linewidth]{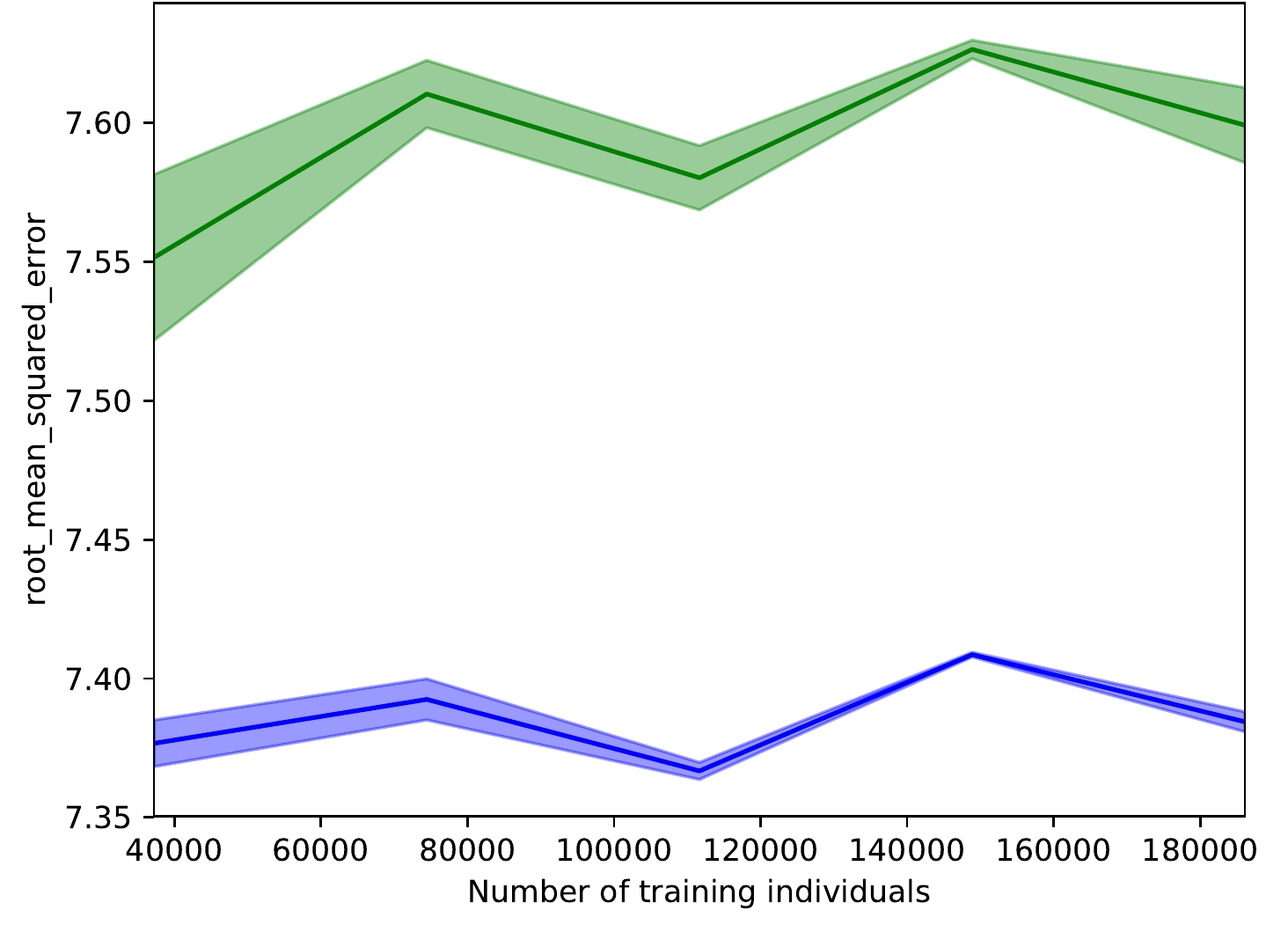}
	\includegraphics[width=0.2\textwidth, trim=0 -19em 0 0]{./fig/model_legend_short}
	\vspace{-1.5ex}
	\caption{Prediction RMSE over number of training individuals for a less informative prior on $\lambda$ and a completely uniformative (\ie uniform) one on $\pi$, $u_0 = [60, 2, 1, 1]$.}
	\label{fig:prior_2}
\end{figure}

We showcase the prediction RMSE results on day 0 of the next cycle for both priors in Figures~\ref{fig:prior_1} and ~\ref{fig:prior_2}, where the blue line represents results for $s\geq 0$ and the green line represents results for $s=0$. Note that these results look similar in magnitude and spread as the prior we have chosen, and we therefore conclude that our method is stable to different choices of priors. 

\subsection*{Performance stability across different dataset sizes and ordering of cycles}
\begin{figure*}[!h]
	\centering
	\includegraphics[width=0.3\textwidth]{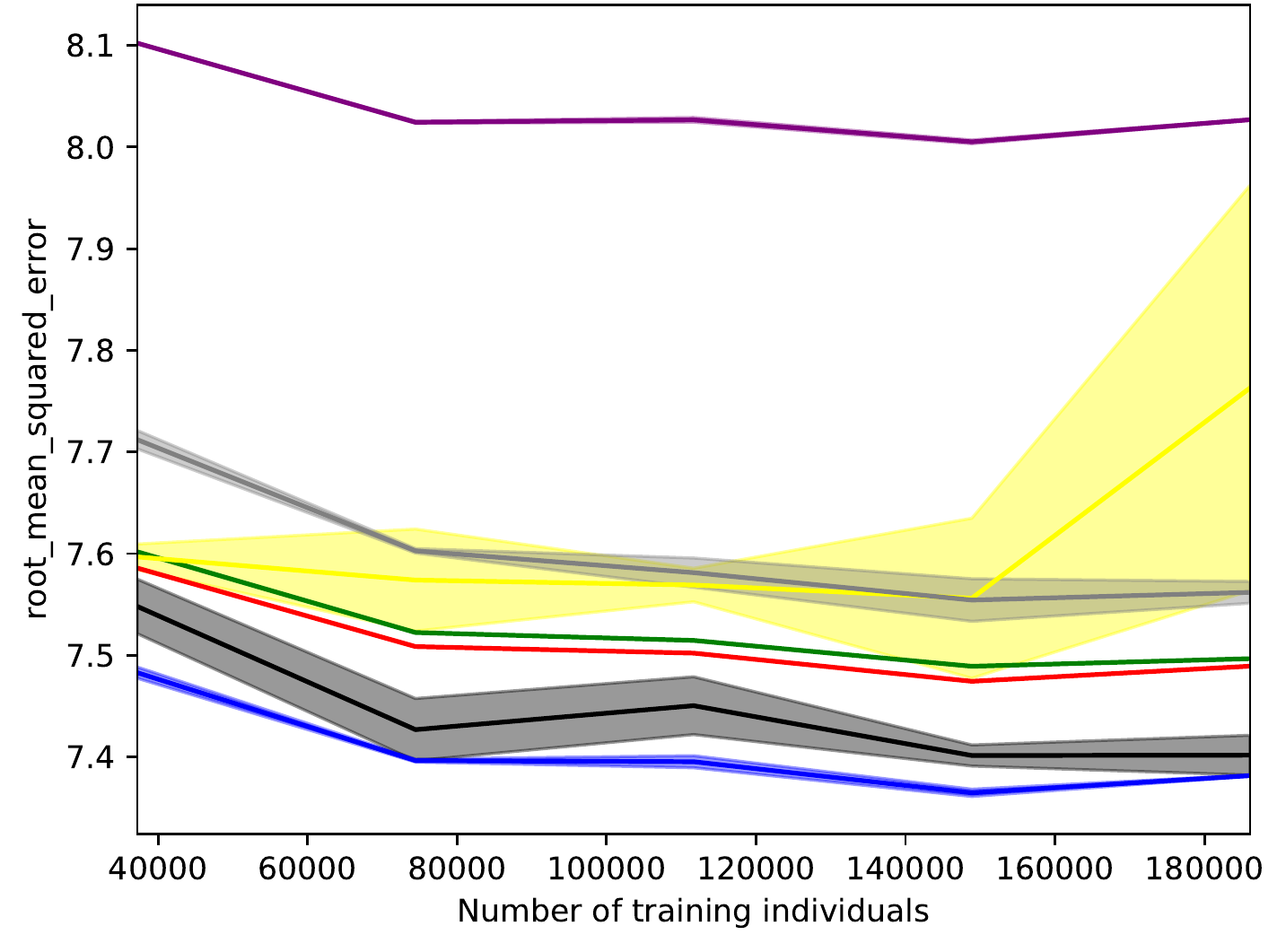}
	\includegraphics[width=0.3\textwidth]{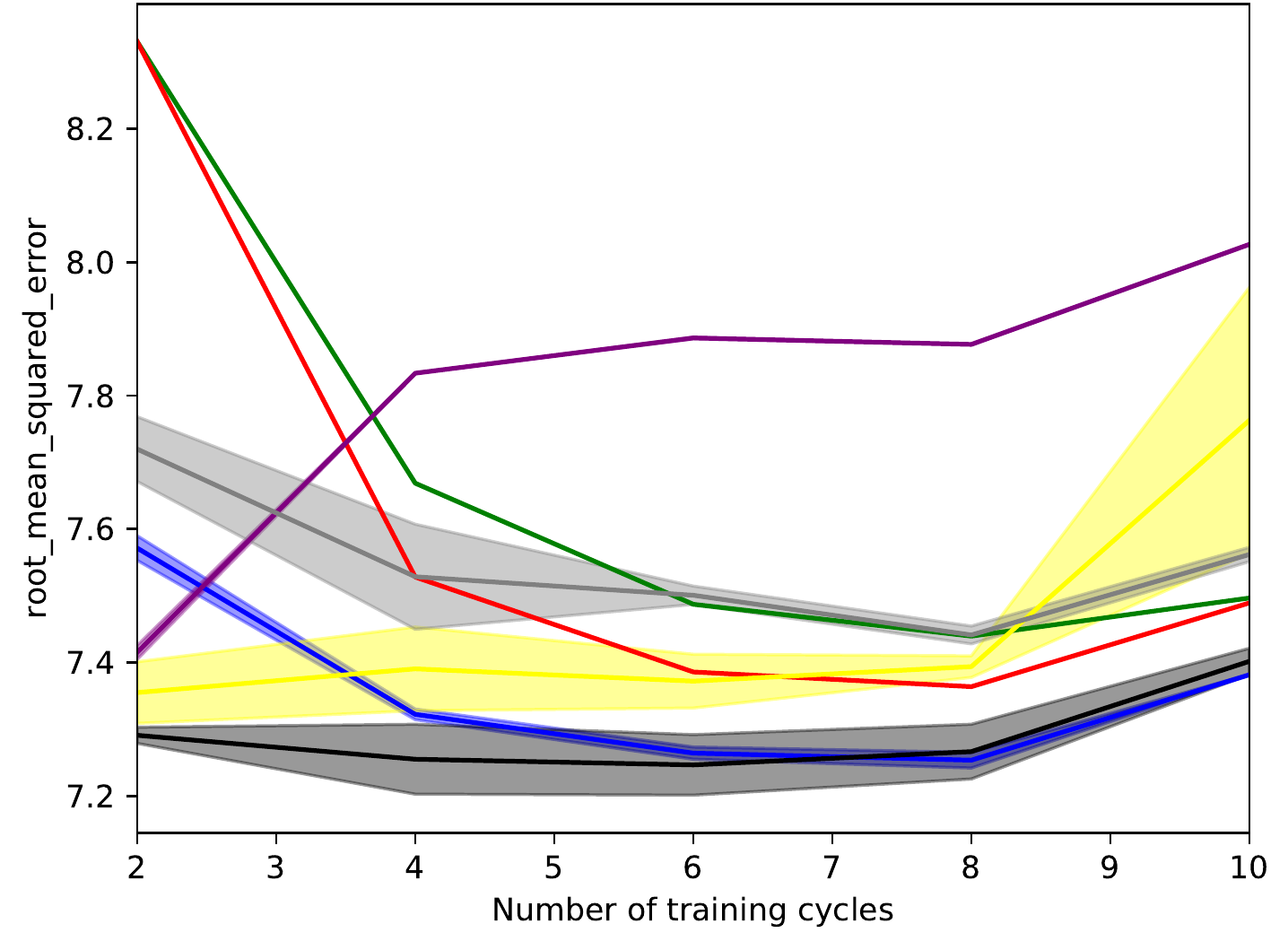}
	\includegraphics[width=0.15\textwidth, trim=0 -12em 0 0]{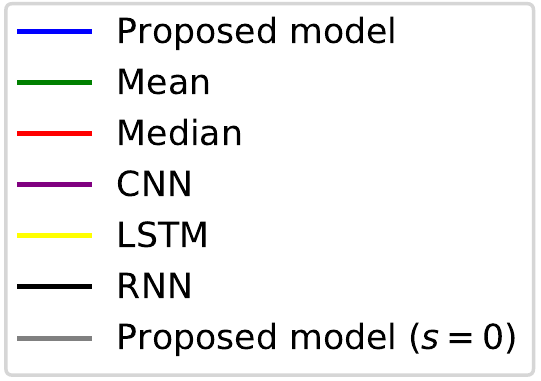}
	\caption{Prediction RMSE for proposed model and baselines on day 0 over number of individuals, $I$ (\textbf{a}) and number of training cycles, $C$ (on the full set of $I$) (\textbf{b}). $C=2$ means $2$ input cycles were used to predict the third and so on.
		\textbf{(a)} Our model outperforms summary statistic-based and neural network-based baselines on day 0 when we account for skipped cycles (blue line), across all subsets of $I$. In addition, our model produces sharper estimates (lower variance) and is stable across $I$ -- with less than $40,000$ users, we have an RMSE less than $7.5$. 
		\textbf{(b)} Our model is robust to different $C$, as shown by consistent RMSE with at least 4 training cycles. Note that all models experience some fluctuations in RMSE depending on number of training cycles; this is due to data randomness, see Figure~\ref{fig:over_C}. 
	}
	\label{fig:pred_by_day_I_C}
\end{figure*}

To demonstrate our model's robustness across different dataset sizes, we showcase prediction RMSE results across different numbers of individuals, $I$ (left) and training cycles, $C$ (right) in Figure~\ref{fig:pred_by_day_I_C}. We see that our model performance is robust to different $I$ and $C$ values -- our model's prediction RMSE remains around 7.5 even with relatively small $I$ or $C$. 

\begin{figure}[h!]
	\centering
	\centering
	\includegraphics[width=0.4\linewidth]{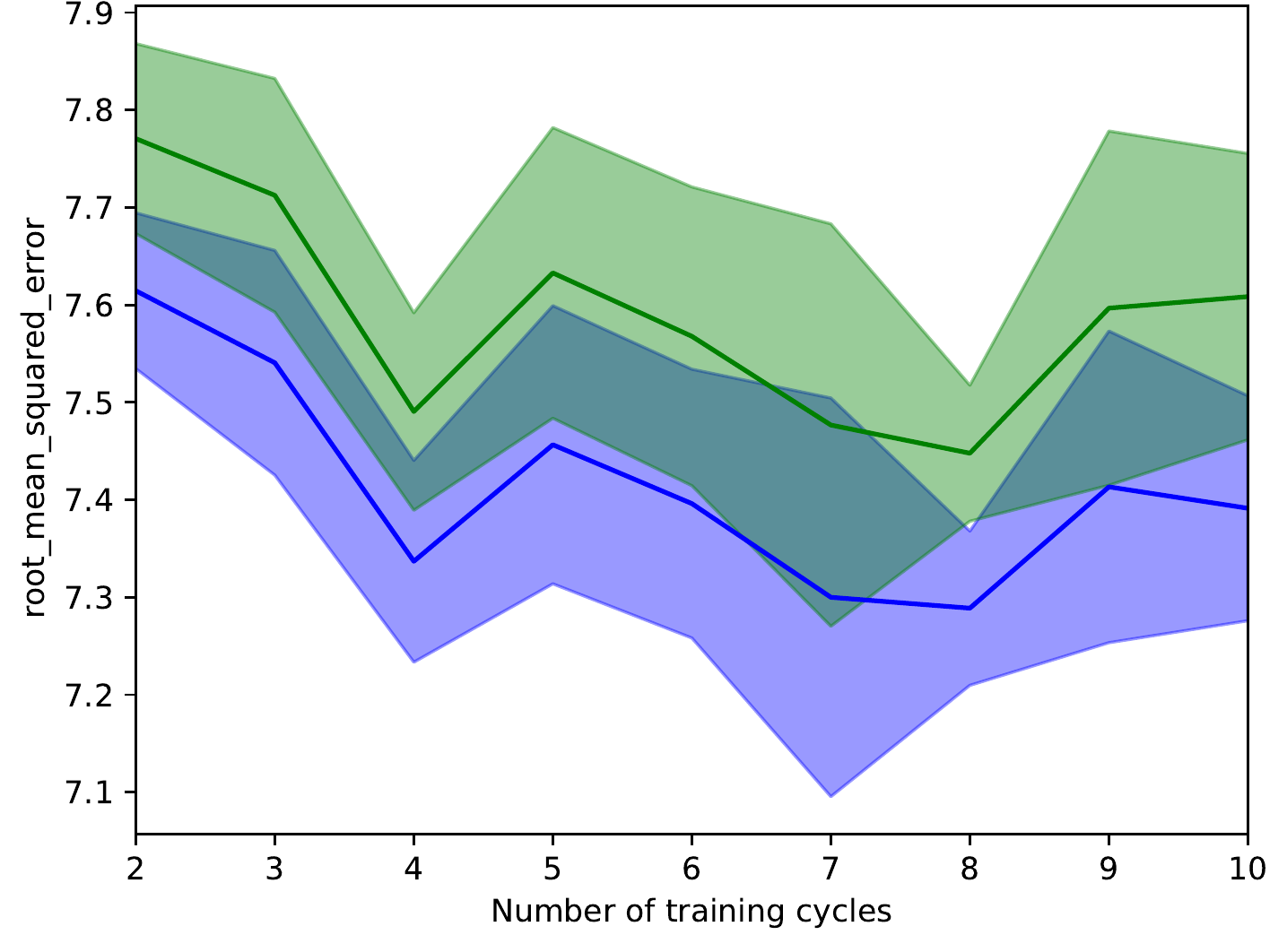}
	\includegraphics[width=0.2\textwidth, trim=0 -19.5em 0 0]{./fig/model_legend_short}
	\vspace{-1.5ex}
	\caption{Prediction RMSE over number of training cycles, averaged over 10 runs of different randomly-drawn datasets of $I=10,000$ users.}
	\label{fig:over_C}
\end{figure}

While our model performance is generally stable to dataset size as in Figure~\ref{fig:pred_by_day_I_C}, we note also that there is some very small magnitude fluctuation in performance with $C=10$. This is due to data randomness -- that is, since we utilize the first $C$ cycles in each training subset, there may be users who happened to have less adherent tracking near the end of their tracking history (i.e., with $C=10$), resulting in a small uptick in prediction RMSE. To showcase this, we perform an experiment utilizing $I=10,000$ users across $10$ runs of our model; for each run, we randomly draw $I=10,000$ users from the full dataset, train our model, and compute predictions. The results of this experiment averaged over the $10$ runs are shown in Figure~\ref{fig:over_C}, where we see that there is some fluctuation in prediction RMSE across $C$ (not just for $C=10$), verifying that the small fluctuation for $C=10$ on the full dataset is an artifact of data randomness. 

\begin{figure}[h!]
	\centering
	\centering
	\includegraphics[width=0.4\linewidth]{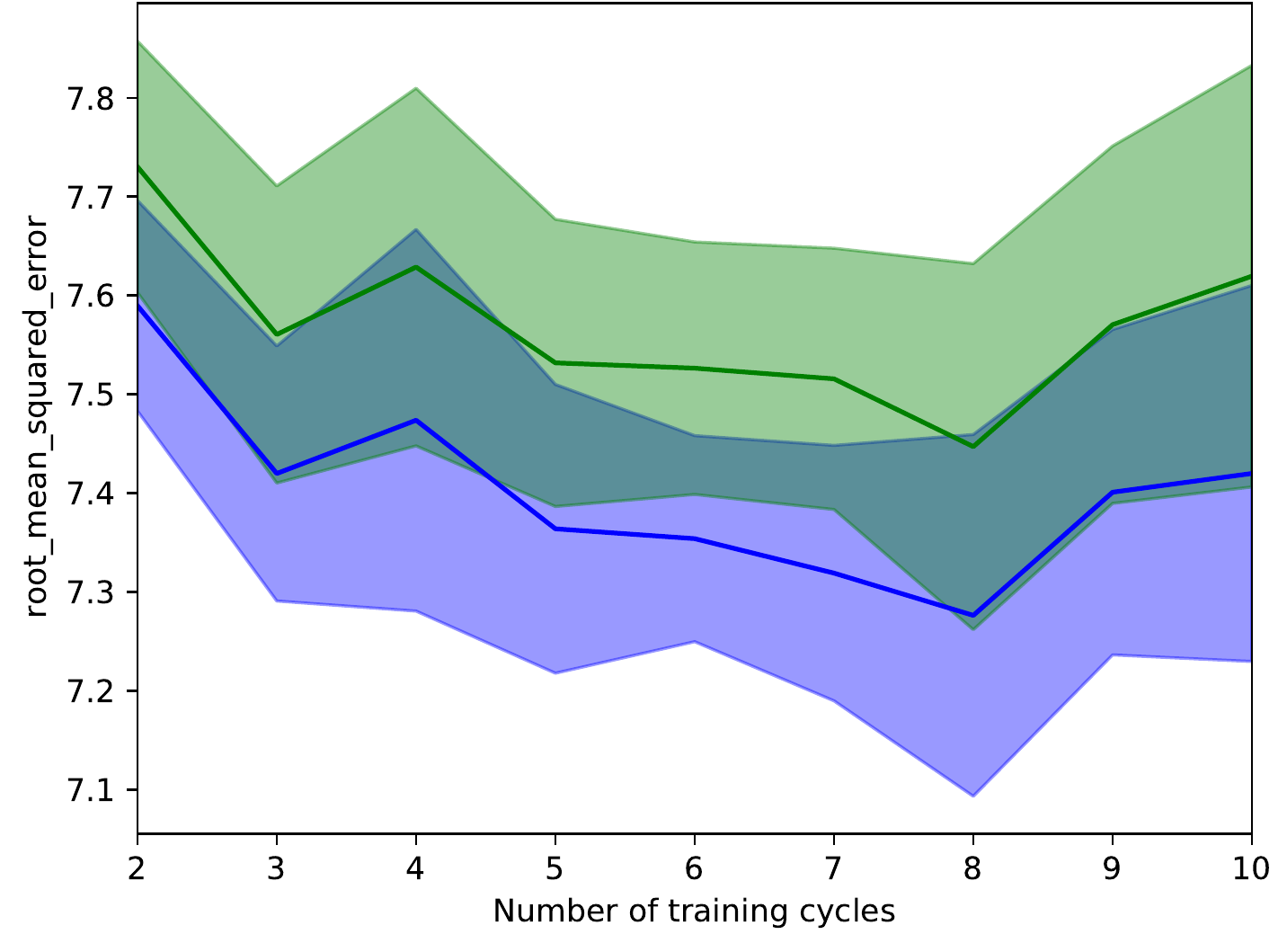}
	\includegraphics[width=0.2\textwidth, trim=0 -19.5em 0 0]{./fig/model_legend_short}
	\vspace{-1.5ex}
	\caption{Prediction RMSE over number of training cycles, averaged over 10 runs of different randomly-drawn datasets of $I=10,000$ users. Here, before we take the first $C$ cycles from each user, we randomly shuffle them.}
	\label{fig:over_C_shuffled}
\end{figure}

To further test the dependency of model predictive performance on the ordering of the observed training cycles, we also run the same experiment with a random shuffling of a user's cycle history before selecting the first $C$ cycles for training. We showcase these results in Figure~\ref{fig:over_C_shuffled} and see again that there are small fluctuations in performance across $C$, verifying further the impact of data randomness. This also showcases the negligible effect of choosing to either take the first $C$ cycles without shuffling (as in Figure~\ref{fig:over_C}) or with shuffling (as in Figure~\ref{fig:over_C_shuffled}).
 
\subsection*{Baseline results with different neural network settings}
\begin{figure}[h!]
	\centering
	\centering
	\includegraphics[width=0.8\linewidth]{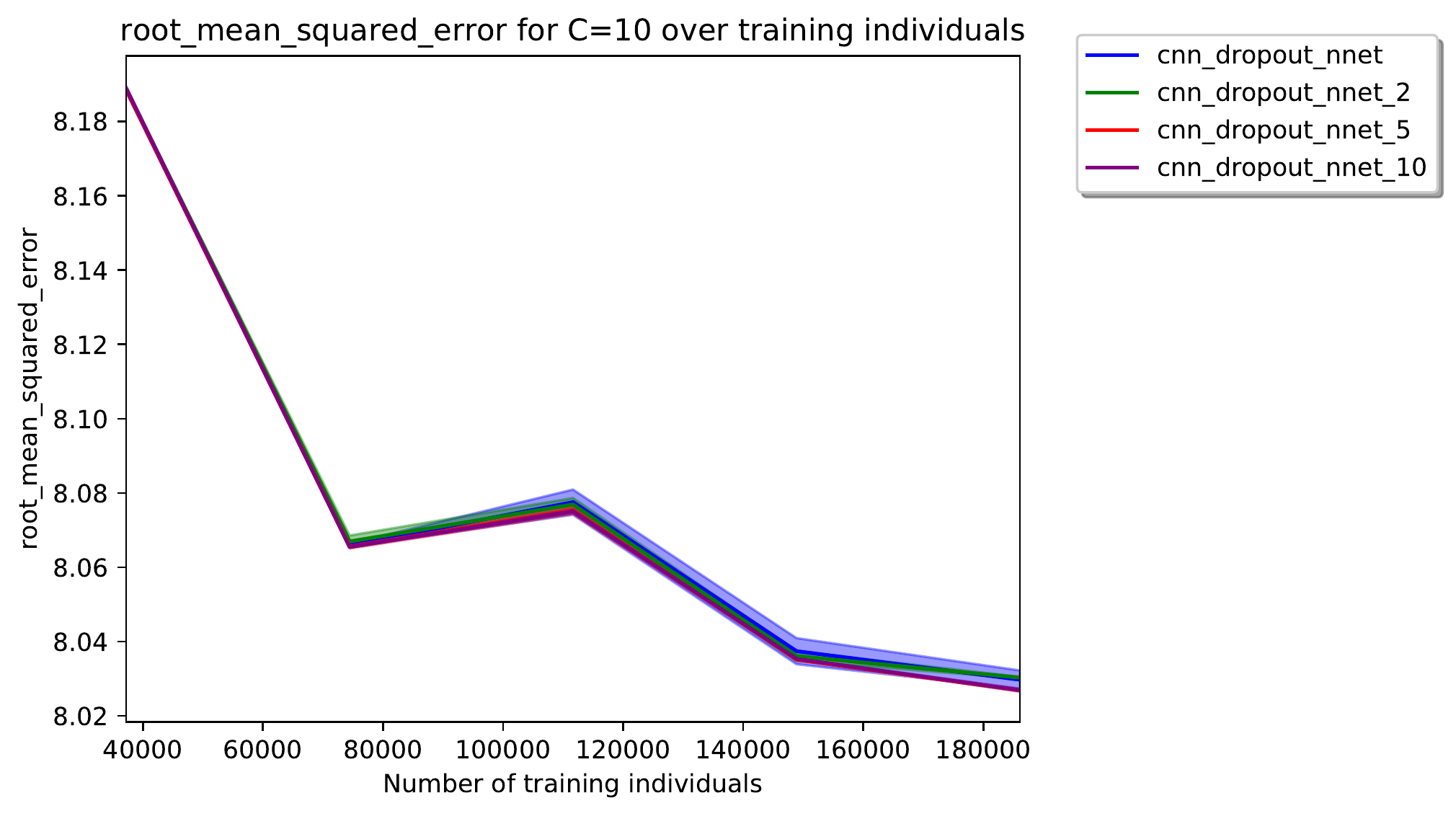}
	\vspace{-1.5ex}
	\caption{Prediction RMSE over number of individuals for CNNs with 1, 2, 5, and 10 layers (blue, green, red, and purple lines, respectively) and a kernel size of $3$.}
	\label{fig:cnn}
\end{figure}

\begin{figure}[h!]
	\centering
	\centering
	\includegraphics[width=0.8\linewidth]{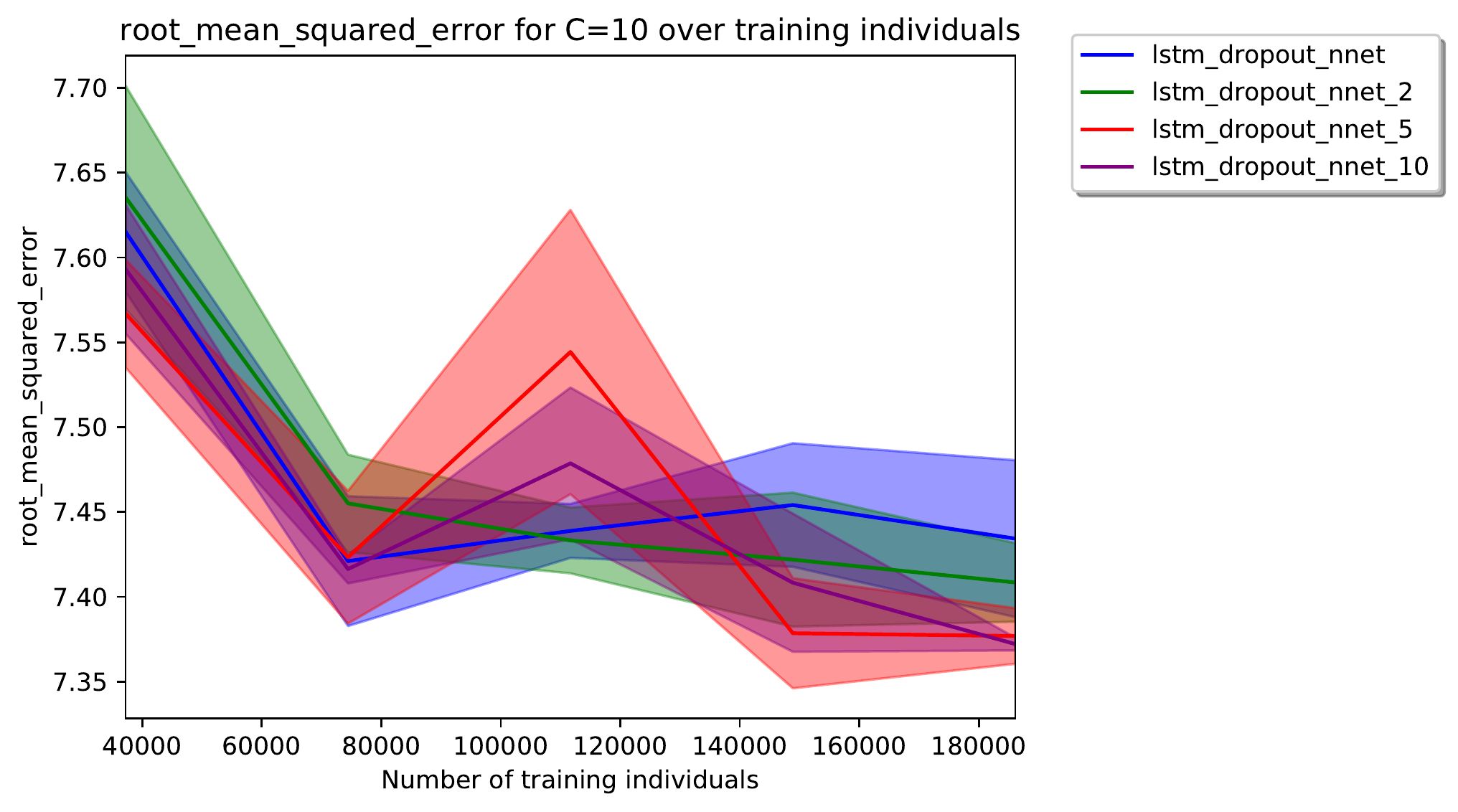}
	\vspace{-1.5ex}
	\caption{Prediction RMSE over number of individuals for LSTMs with 1, 2, 5, and 10 layers (blue, green, red, and purple lines, respectively) and a hidden size of $3$.}
	\label{fig:lstm}
\end{figure}

\begin{figure}[h!]
	\centering
	\centering
	\includegraphics[width=0.8\linewidth]{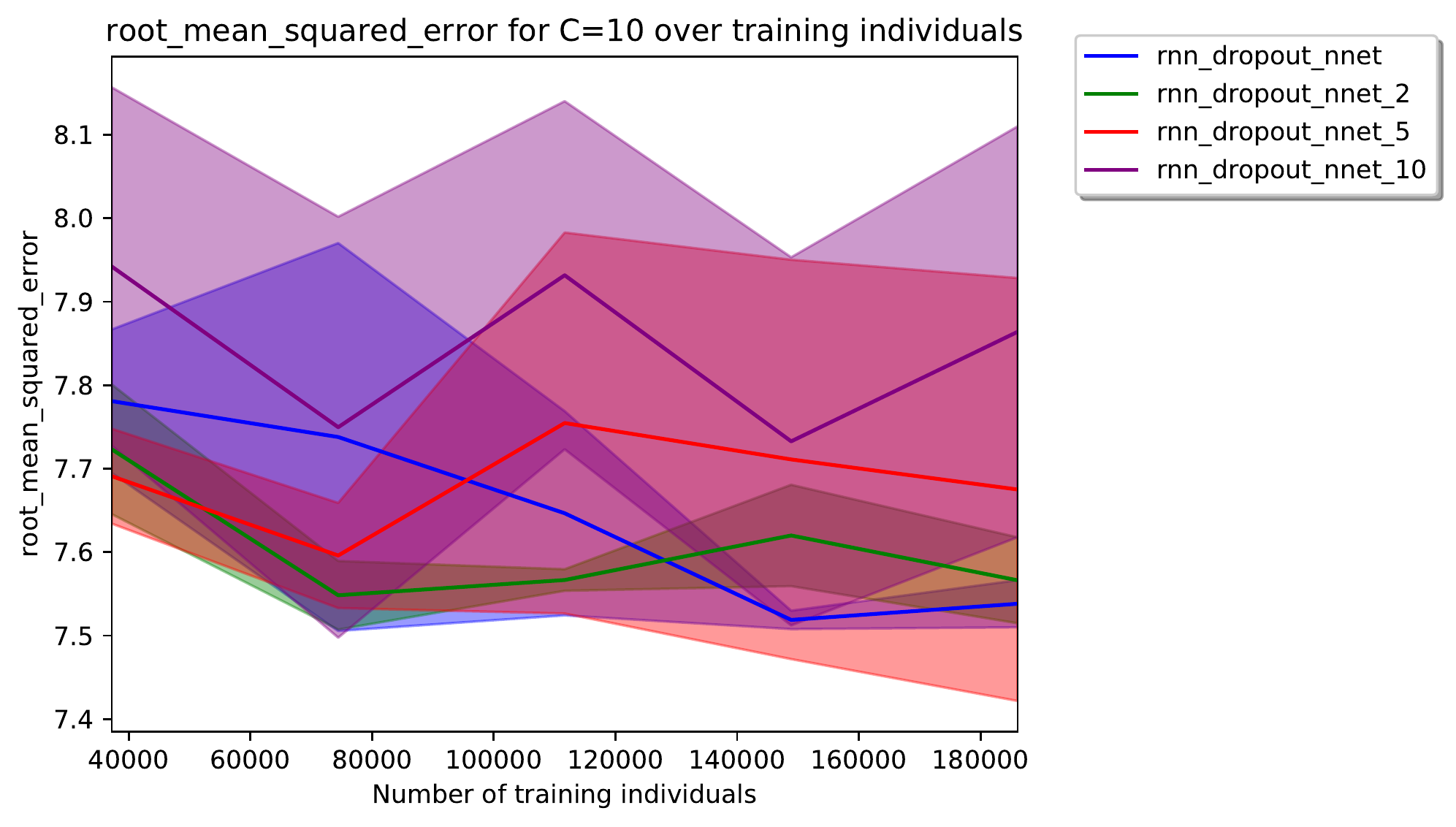}
	\vspace{-1.5ex}
	\caption{Prediction RMSE over number of individuals for RNNs with 1, 2, 5, and 10 layers (blue, green, red, and purple lines, respectively) and a hidden size of $3$.}
	\label{fig:rnn}
\end{figure}

In the results of our main manuscript, we utilize neural network-based baselines with one layer and a kernel size or hidden size of $3$. To assess the performance of neural network-based baselines with different settings, we test ($i$) different numbers of layers and ($ii$) different kernel and hidden sizes (using a kernel or hidden size equal to the number of training cycles $C$ instead of fixed at $3$). Figures~\ref{fig:cnn},~\ref{fig:lstm}, and~\ref{fig:rnn} showcase the performance RMSEs across $I$ for 1, 2, 5, and 10-layer CNNs, LSTMs, and RNNs, respectively (with fixed kernel or hidden size of $3$). Figures~\ref{fig:cnn}, ~\ref{fig:lstm}, and ~\ref{fig:rnn} showcase the performance RMSEs across $I$ for 1, 2, 5, and 10-layer CNNs, LSTMs, and RNNs, respectively (with kernel or hidden size of $C=10$). We see that across the number of layers and kernel or hidden size of $3$ or $C=10$, the prediction RMSE is stable, with average differences of at most $0.5$ between different settings. Therefore, we conclude that one-layer neural networks, with fixed kernel or hidden size of $3$, are reasonable baselines.

\begin{figure}[h!]
	\centering
	\centering
	\includegraphics[width=0.8\linewidth]{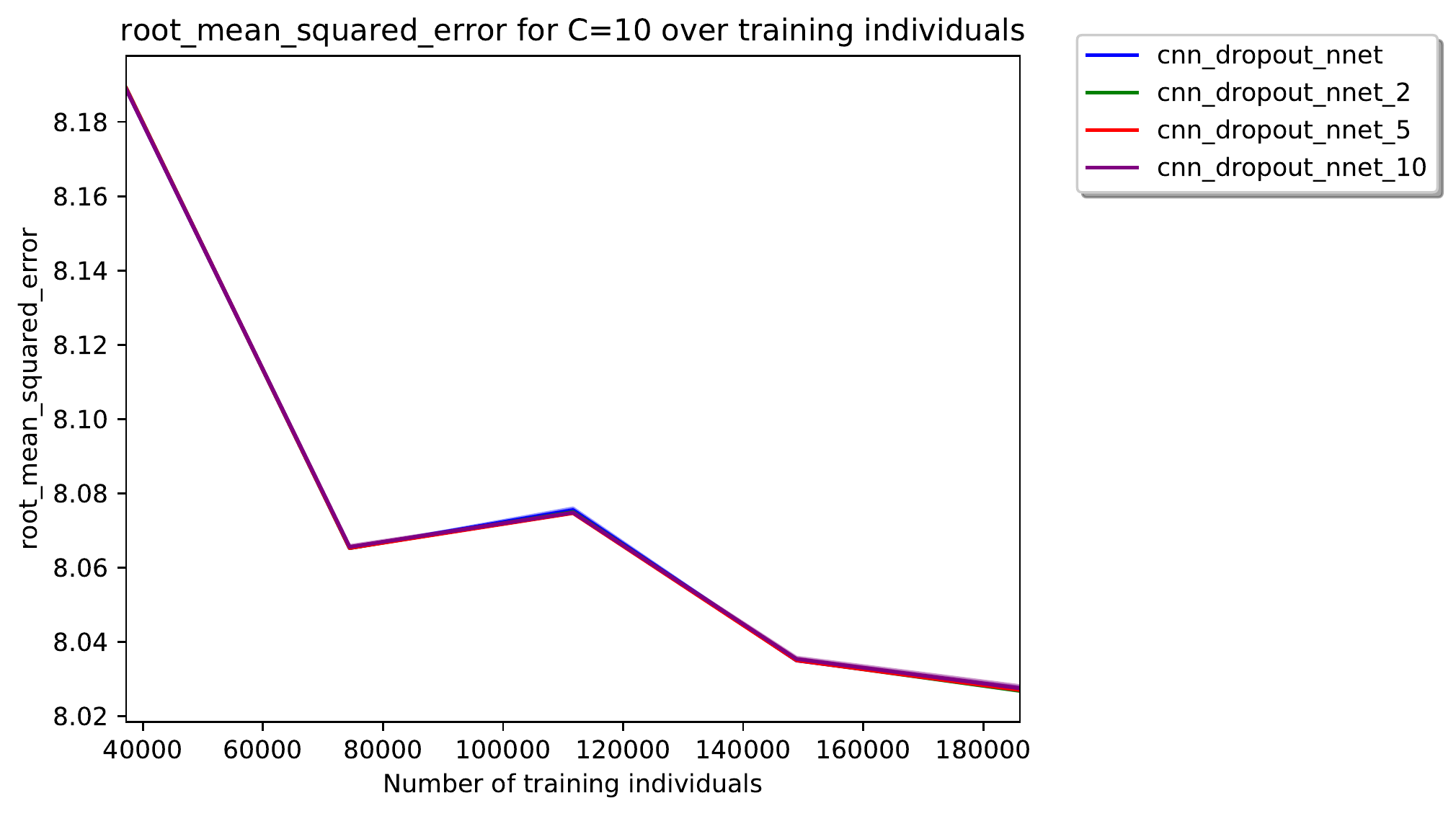}
	\vspace{-1.5ex}
	\caption{Prediction RMSE over number of individuals for CNNs with 1, 2, 5, and 10 layers (blue, green, red, and purple lines, respectively) and a kernel size of $C=10$.}
	\label{fig:cnn_C}
\end{figure}

\begin{figure}[h!]
	\centering
	\centering
	\includegraphics[width=0.8\linewidth]{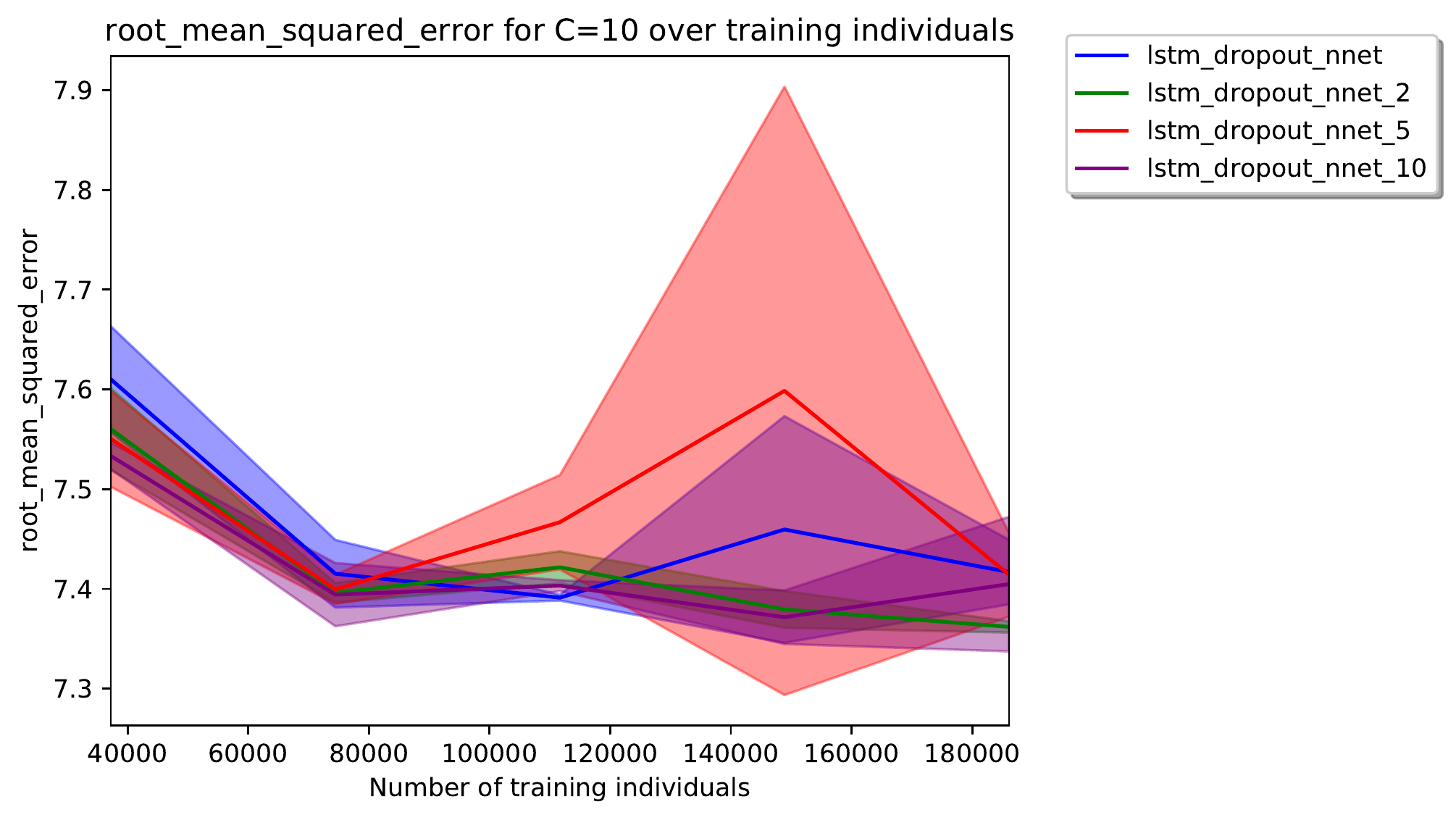}
	\vspace{-1.5ex}
	\caption{Prediction RMSE over number of individuals for LSTMs with 1, 2, 5, and 10 layers (blue, green, red, and purple lines, respectively) and a hidden size of $C=10$.}
	\label{fig:lstm_C}
\end{figure}

\begin{figure}[h!]
	\centering
	\centering
	\includegraphics[width=0.8\linewidth]{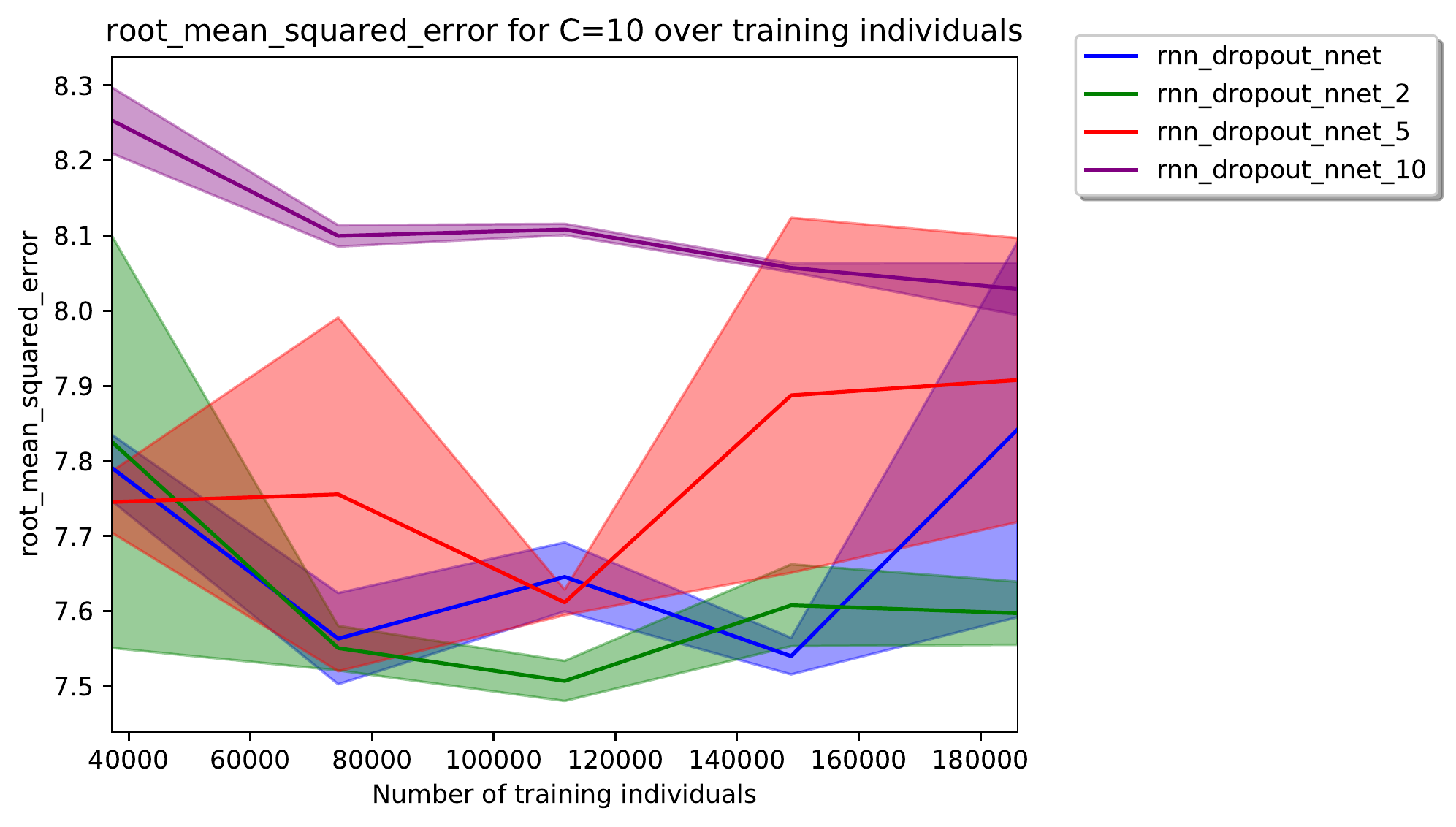}
	\vspace{-1.5ex}
	\caption{Prediction RMSE over number of individuals for RNNs with 1, 2, 5, and 10 layers (blue, green, red, and purple lines, respectively) and a hidden size of $C=10$.}
	\label{fig:rnn_C}
\end{figure}
\end{document}